\documentclass[10pt,twocolumn,letterpaper]{article}

\usepackage{iccv}              

\usepackage{multirow} 
\usepackage{utfsym}

\definecolor{iccvblue}{rgb}{0.21,0.49,0.74}
\usepackage[pagebackref,breaklinks,colorlinks,allcolors=iccvblue]{hyperref}

\definecolor{barrier}{RGB}{112,128,144}

\title{MuDG: Taming \underline{Mu}lti-modal \underline{D}iffusion with \underline{G}aussian Splatting \\for {U}rban Scene {R}econstruction}

\author{
    Yingshuang Zou\textsuperscript{\rm 1,2}\thanks{Equal contribution.}  
    ~Yikang Ding\textsuperscript{\rm 2}$^{\ast}$\thanks{Project Leader.}~ 
    ~Chuanrui Zhang\textsuperscript{\rm 1,2}
    ~Jiazhe Guo\textsuperscript{\rm 1,2}
    ~Bohan Li\textsuperscript{\rm 2,4}
    \\
    Xiaoyang Lyu\textsuperscript{\rm 5}  
    ~Feiyang Tan\textsuperscript{\rm 3} 
    ~Xiaojuan Qi\textsuperscript{\rm 5} 
    ~Haoqian Wang\textsuperscript{\rm 1}\thanks{Corresponding author.} \vspace{0.2cm}\\
    $^{1}$THU~ $^{2}$MEGVII~ $^{3}$Mach Drive~
    $^{4}$SJTU~ $^{5}$HKU \vspace{0.2cm}
    \\
    \url{https://github.com/heiheishuang/MuDG}
}

\begin{document}
\twocolumn[{%
    \renewcommand\twocolumn[1][]{#1}%
    \maketitle
    \centering
    \vspace{-16pt}
    \includegraphics[width=0.93\textwidth]{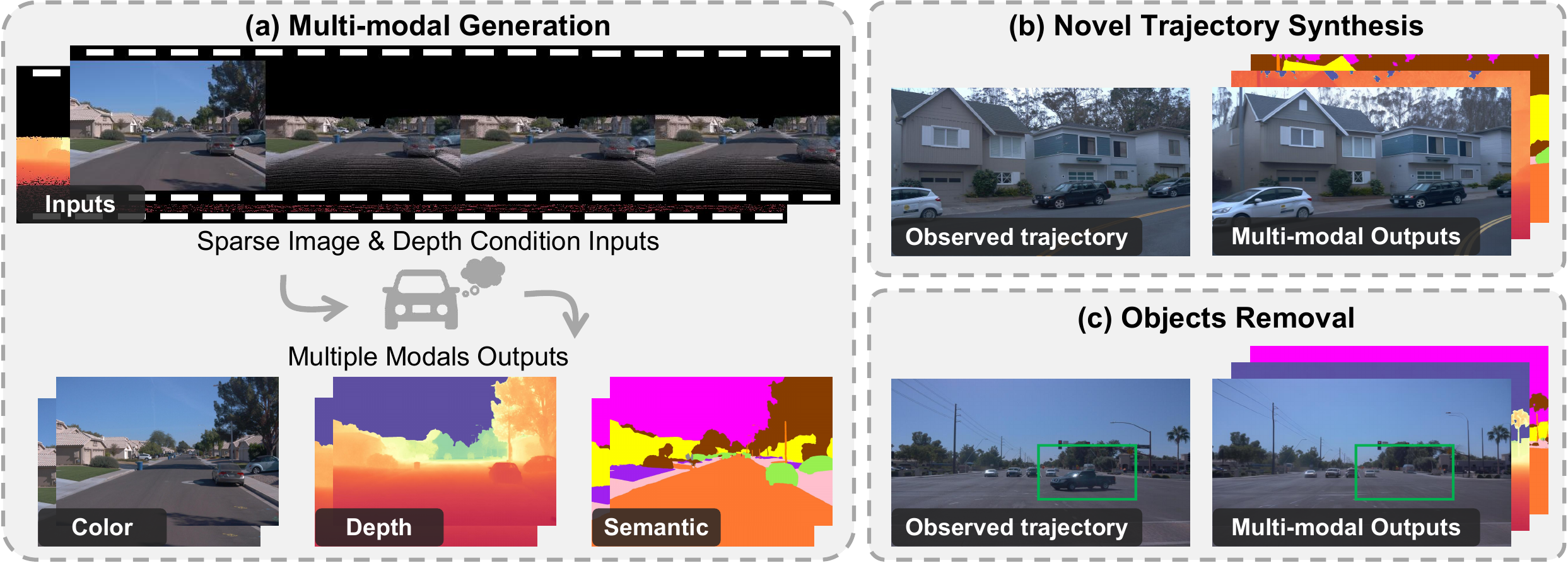}
    \vspace{-6pt}
    \captionof{figure}
    {
    Given sparse sequential image-depth inputs, our multi-modal diffusion model enables controllable novel view synthesis, abstaining per-scene optimization. It also serves as a supervised signal to enhance the Gaussian Splatting model. Furthermore, its strong controllability allows for scene editing (e.g., object removal, background modification), providing valuable data for autonomous driving.
    }
    \label{fig:teaser}
    \vspace{10pt}
}]
{\let\thefootnote\relax\footnotetext{{ $^{\ast}$ Equal contribution.  $\dag$ Project Leader. $\ddagger$ Corresponding author.}}}

\vspace{-8pt}
\begin{abstract}
Recent breakthroughs in radiance fields have significantly advanced 3D scene reconstruction and novel view synthesis (NVS) in autonomous driving. Nevertheless, critical limitations persist: reconstruction-based methods exhibit substantial performance deterioration under significant viewpoint deviations from training trajectories, while generation-based techniques struggle with temporal coherence and precise scene controllability. To overcome these challenges, we present \textbf{MuDG}, an innovative framework that integrates \textbf{Mu}lti-modal \textbf{D}iffusion model with \textbf{G}aussian Splatting (GS) for {U}rban Scene {R}econstruction. MuDG leverages aggregated LiDAR point clouds with RGB and geometric priors to condition a multi-modal video diffusion model, synthesizing photorealistic RGB, depth, and semantic outputs for novel viewpoints. This synthesis pipeline enables feed-forward NVS without computationally intensive per-scene optimization, providing comprehensive supervision signals to refine 3DGS representations for rendering robustness enhancement under extreme viewpoint changes. Experiments on the Open Waymo Dataset demonstrate that MuDG outperforms existing methods in both reconstruction and synthesis quality.
\end{abstract}
\vspace{-10pt}    
\section{Introduction}
\label{sec:intro}

With the advancement of radiance field-based reconstruction techniques (\eg, NeRF~\cite{mildenhall2021nerf, zhang2020nerf++, fridovich2022plenoxels}, 3D Gaussian Splatting~\cite{kerbl20233dgs, yu2024mip, huang20242dgs, zhang2024transplat}), 3D reconstruction and novel view synthesis (NVS) have been widely applied in AR/VR, robotics, and many other fields~\cite{long2024wonder3d, fu2024geowizard}.
In autonomous driving, such techniques enable dynamic urban scene reconstruction and view synthesis~\cite{chen2023PVG, yang2023emernerf, huang2024S3Gaussian, yan2024streetgs, zou2024m, zhou2024hugs, yang2023unisim}, offering a promising solution for synthetic data generation and closed-loop simulation.
However, reconstructing and rendering high-fidelity sensor observations (RGB videos, LiDAR point clouds, and dense depth maps) remains challenging in complex driving environments.\looseness=-1

To address this challenge, reconstruction-based approaches
~\cite{unisim, yang2023emernerf, chen2023PVG, huang2024S3Gaussian, yan2024streetgs, zhou2024hugs, zhou2024drivinggaussian} extend NeRF and 3DGS to dynamic urban environments. These methods achieve scene decomposition by separating dynamic objects from static backgrounds via bbox tracking~\cite{unisim, yan2024streetgs, zhou2024hugs, zhou2024drivinggaussian} or self-supervised decomposition~\cite{chen2023PVG, huang2024S3Gaussian, yang2023emernerf}, demonstrating strong reconstruction quality and rendering performance along captured trajectories.

However, these reconstruction-centric methods still face limitations in NVS. When rendering viewpoints deviate significantly from the original trajectories, visual quality deteriorates rapidly, with severe artifacts emerging due to insufficient scene reconstruction. This fundamental limitation constrains their applicability in simulation scenarios requiring flexible viewpoint control.
Meanwhile, emerging generation-based methods leverage diffusion models~\cite{gao2023magicdrive, li2024uniscene, wang2023drivewm, wang2023drivedreamer, zhao2024drivedreamer2, yu2024viewcrafter} conditioned on control signals (\eg, Bird’s-Eye-View (BEV) layouts and waypoints) for urban scene synthesis. While these generative approaches produce high-fidelity outputs, they struggle with consistency across multiple generations. The inherent stochasticity of diffusion processes introduces temporal incoherence and trajectory variation artifacts, which pose challenges for deployment in simulation systems and corner-case generation scenarios.
We attribute the performance degradation of reconstruction-based methods under out-of-trajectory viewpoints to insufficient supervision during scene optimization, while the inconsistency in generative approaches arises from the lack of reliable and detailed conditions.

To address these challenges, we propose MuDG, a framework that integrates a controllable \textbf{Mu}lti-modal \textbf{D}iffusion model with \textbf{G}aussian Splatting (GS) for {U}rban Scene {R}econstruction.
At its core, MuDG features a multi-modal video generation model conditioned on sparse point cloud projections, synthesizing RGB, depth, and semantic videos from novel viewpoints, as shown in Fig.~\ref{fig:teaser}.
The generated results can serve not only as rendering outputs for novel views without requiring per-scene optimization but also as supervisory signals for training Gaussian Splatting models, enabling robust rendering under large viewpoint changes.
Specifically, we aggregate LiDAR point clouds across multiple frames and separate dynamic and static elements using tracked bounding boxes. These point clouds retain RGB and geometry information from their original trajectories.
For novel view synthesis, our model takes sparse RGB and depth from aggregated point cloud projections as input and generates dense results aligned with the projected viewpoints.\looseness=-1

This multi-modal design leverages video generation priors to simulate GS rendering in a feed-forward manner, allowing direct application to novel scenes without per-scene optimization.
Furthermore, the synthesized multi-model results provide rich supervision for subsequent GS training, which is important in reconstructing dynamic urban scenes, enabling the enhanced GS models to maintain high rendering quality even under substantial camera movements. This effectively mitigates the performance degradation observed in conventional GS systems when extrapolating beyond training viewpoints.
Our contributions are summarized as follows:\looseness=-1

\begin{itemize}

    

    \item A novel controllable multi-modal diffusion model for feed-forward synthesis of high-fidelity RGB, depth, and semantic segmentation outputs in novel view synthesis.
    
    \item Integration of multi-modal generation with Gaussian Splatting, achieving significant improvements in out-of-trajectory NVS quality for urban scenes.

    \item Extensive experiments on the Open Waymo Dataset (WOD) benchmark~\cite{sun2020scalability} indicate our approach outperforms existing methods.\looseness=-1

\end{itemize}

\begin{figure*}[!t]
\centering
\includegraphics[width=\textwidth]{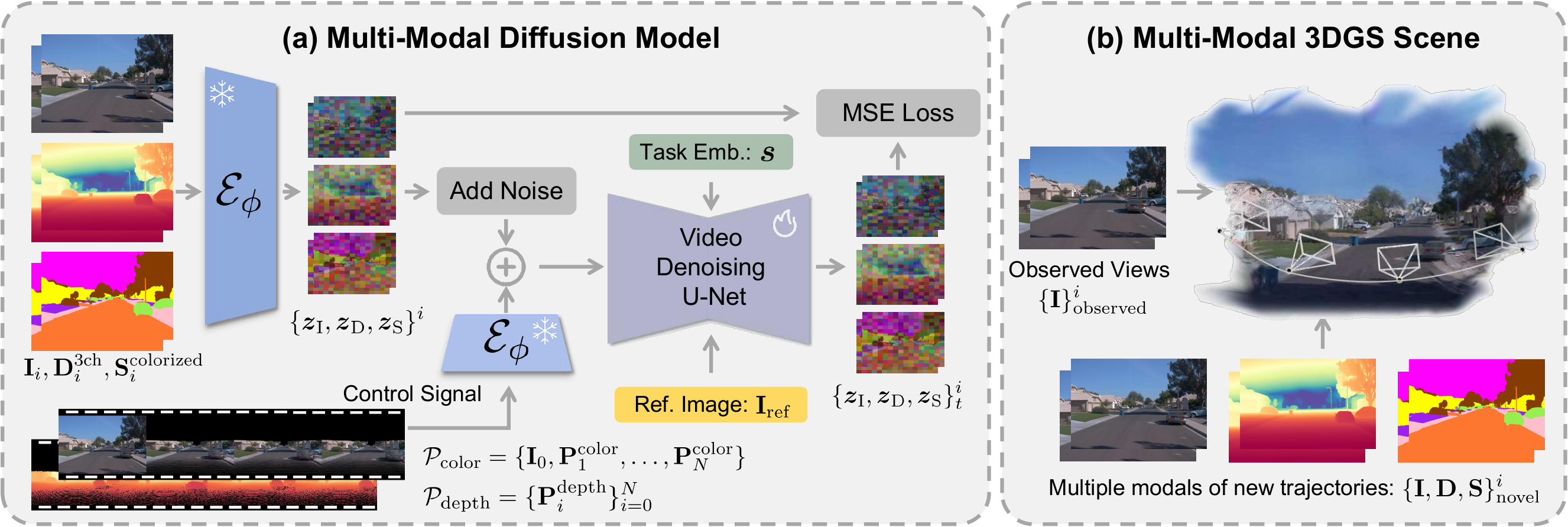} 
\caption{\textbf{Framework of MuDG.} 
(a) Training Phase of the Multi-modal Diffusion Model (MDM). 
Given a reference image and sparse conditions $\mathcal{P}_{\text{color}}$ and $\mathcal{P}_{\text{depth}}$, we fine-tune the Video Diffusion Model to generate a color image $\mathbf{I}_{i}$, a 3-channel depth map $\mathbf{D}_{i}^{\text{3ch}}$ and a colorized semantic map $\mathbf{S}_{i}^{\text{colorized}}$ using multi-task embeddings $\boldsymbol{s}$.
(b) Pipeline of the Multi-modal 3DGS Scene.
Using the dense output from the MDM module, we optimize a 3D Gaussian Splatting (3DGS) representation with better geometry-semantic consistency.
}
\label{fig:pipeline}
\end{figure*}

\section{Related Work}

\subsection{Novel View Synthesis for Urban Scene}

Recent advancements in NeRF~\cite{mildenhall2021nerf, zhang2020nerf++, fridovich2022plenoxels, tonderski2024neurad} and 3D Gaussian Splatting (GS)~\cite{kerbl20233dgs, yu2024mip, huang20242dgs} have driven progress in novel view synthesis for urban scenes~\cite{unisim, yang2023emernerf, chen2023PVG, huang2024S3Gaussian, yan2024streetgs, zhou2024hugs, zhou2024drivinggaussian}, emphasizing real-time performance, spatial-temporal coherence, and controllability. Pioneering works like 3D Gaussian Splatting~\cite{gaussian_splatting} achieve real-time radiance field rendering by optimizing anisotropic 3D Gaussians, enabling high-fidelity novel-view synthesis at 1080p resolution. To address dynamic urban scenes, Periodic Vibration Gaussian (PVG)~\cite{chen2023PVG} introduces temporal dynamics into Gaussian representations, unifying static and dynamic elements without manual annotations, while Street Gaussians~\cite{yan2024streetgs} explicitly model foreground vehicles and backgrounds for efficient rendering.
EmerNeRF~\cite{yang2023emernerf} proposes a self-supervised spatial-temporal decomposition framework, capturing geometry, motion, and semantics without ground-truth annotations, and enhancing 3D perception tasks via lifted 2D foundation model features. 
DriveDreamer4D~\cite{zhao2024drivedreamer4d} bridges 4D reconstruction and world models, leveraging video generation priors to improve spatiotemporal coherence in novel trajectory views, outperforming prior Gaussian-based methods.
STORM~\cite{yang2024storm} adopts a feed-forward Transformer architecture to infer dynamic 3D Gaussians and velocities, enabling efficient large-scale outdoor scene reconstruction. FreeVS~\cite{wang2024freevs} introduces generative view synthesis for free trajectories, bypassing per-scene optimization through pseudo-image priors, while DrivingForward~\cite{tian2024drivingforward} achieves feed-forward reconstruction from sparse surround-view inputs via self-supervised pose and depth estimation.  
Despite these advancements, existing methods struggle to synthesize high-fidelity multi-modal sensor observations (e.g., RGB videos, LiDAR point clouds, dense depth maps) in complex driving environments, particularly when rendering viewpoints that deviate significantly from recorded trajectories. To address these limitations, we propose a controllable multi-modal diffusion model to enable high-quality out-of-trajectory novel view synthesis.\looseness=-1


\subsection{Controllable Video Generation for Urban Scene}

Recent advances in generative models have significantly propelled urban scene generation for autonomous driving, addressing challenges in 3D geometry control, multi-view consistency, and spatiotemporal dynamics~\cite{gao2023magicdrive, li2024uniscene, wang2023drivewm,li2024one,wang2023drivedreamer,li2024hierarchical,zhao2024drivedreamer2, gao2024vista}. MagicDrive~\cite{gao2023magicdrive} introduces a framework for street-view synthesis with precise 3D controls (e.g., camera poses, road maps) via cross-view attention, enhancing 3D object detection and BEV segmentation. Extending this, MagicDrive3D~\cite{gao2024magicdrive3d} proposes a controllable 3D generation pipeline leveraging deformable Gaussian splatting to address unbounded street scenes, enabling high-fidelity any-view rendering. For dynamic scenarios, MagicDriveDiT~\cite{gao2024magicdrivedit} employs a DiT-based architecture with spatial-temporal latent conditioning to generate high-resolution, long-duration driving videos, while DreamDrive~\cite{mao2024dreamdrive} combines video diffusion and hybrid Gaussian representations to synthesize 4D scenes with 3D-consistent dynamic video rendering.
UniScene~\cite{li2024uniscene} presents an occupancy-centric approach to unify semantic, visual, and LiDAR data generation, reducing layout-to-data complexity through hierarchical learning strategies.
DriveDreamer-2~\cite{zhao2024drivedreamer2} further enhances customization by integrating LLMs to generate user-defined driving videos with improved temporal coherence.
Stag-1~\cite{wang2024stag} advances 4D simulation by decoupling spatial-temporal dynamics and leveraging point cloud reconstruction for photo-realistic, viewpoint-agnostic scene evolution.
{Recent concurrent works, such as StreetCrafter~\cite{yan2024streetcrafter} and FreeVS~\cite{wang2024freevs}, integrate generative diffusion models for controllable novel-view synthesis. However, these methods overlook the critical role of multi-modal data in achieving high-quality reconstructions, resulting in suboptimal performance.
In this work, we propose a novel controllable multi-modal diffusion model that simultaneously synthesizes high-quality RGB, depth, and semantic outputs for feed-forward novel-view synthesis, outperforming existing methods in terms of synthesis quality.}\looseness=-1

\vspace{-0.2cm}
\section{Method}
In this section, we introduce \textbf{MuDG}, a framework that integrates a controllable \textbf{Mu}lti-modal \textbf{D}iffusion model with \textbf{G}aussian Splatting (GS) for Urban Scene \textbf{R}econstruction.

\subsection{Overview}
As illustrated in Fig.~\ref{fig:pipeline}, our framework comprises two core components: a multi-modal diffusion model (MDM) and a Gaussian Splatting (GS) module. 
To enable feed-forward multi-modal synthesis, our MDM employs sparse RGB-depth conditions derived from fused LiDAR point clouds. We first separate dynamic objects from static backgrounds using tracking bounding boxes to create fused LiDAR point clouds. These point clouds are subsequently reprojected into sparse color and depth maps through perspective projection, forming the conditional inputs for our model.

As detailed in Sec.~\ref{subsec:multi-modal}, MDM learns dense reconstructions (RGB, depth, and semantics) through joint training on sparse input pairs and ground-truth multi-modal data. By manipulating camera extrinsics, we generate novel conditional inputs through virtual viewpoint projection of the fused point cloud. This enables the diffusion model to synthesize reconstructed outputs ($I_{i}^{\text{recon}}, D_{i}^{\text{recon}}, S_{i}^{\text{recon}}$) for arbitrary perspectives through iterative denoising.

The GS module in Sec.~\ref{subsec:gs} is then optimized using the reconstructed virtual viewpoints as supervision, minimizing photometric and geometric discrepancies between rendered and reconstructed views. This optimization strategy ensures scene coherence under extreme camera motions while maintaining multi-modal alignment across viewpoints.

\subsection{Controllable Multi-modal Diffusion Model}\label{subsec:multi-modal}

\subsubsection{Preliminaries}
Video diffusion models have recently achieved significant progress, primarily consisting of two key processes to learn the data distribution~\cite{ho2020denoising, blattmann2023stable}.
In the forward diffusion process, at each time step $t \in \{1, \dots, T\}$, a gaussian noise $\boldsymbol{\epsilon} \sim \mathcal{N}(0, I)$ is sampled and added to the initial latent representation $\mathbf{x}_0$ to obtain the noisy latent $x_t$. This process can be formally expressed as:
\vspace{-0.1cm}
\begin{equation}
\boldsymbol{x}_t = \sqrt{\overline{\alpha}_t} \boldsymbol{x}_0 + \sqrt{1 - \overline{\alpha}_t} \boldsymbol{\epsilon},
\end{equation}
\vspace{-0.1cm}
where $\boldsymbol{x}_t$ denotes the noisy latent at diffusion step $t$, and $\overline{\alpha}_t$ is the noise scheduling parameter.

In the reverse process, a denoising network $\mathcal{F}_\theta$, parameterized by $\theta$, is used to iteratively remove noise from $\boldsymbol{x}_t $ and recover $\boldsymbol{x}_{t-1}$. During training, the network parameters $\theta$ are optimized by minimizing the following loss function:
\vspace{-0.1cm}
\begin{equation}
\mathcal{L} = \mathbb{E}_{\boldsymbol{x}_0, \boldsymbol{c}, t \sim \mathcal{U}(T), \boldsymbol{\epsilon} \sim \mathcal{N}(0, I)}  \| \epsilon - \mathcal{F}_\theta(\boldsymbol{x}_t, \boldsymbol{c}, t) \|_2^2,
\end{equation}
\vspace{-0.1cm}
where $\boldsymbol{c}$ represents optional conditioning information. At inference time, a noise sample $\boldsymbol{x}_{T}$ is first drawn from a standard Gaussian distribution, and the denoising network $\mathcal{F}_\theta$ is applied iteratively to reconstruct the final output, such as an image or video frame.

\subsubsection{Multi-modal Latent Encoding and Decoding}
Following the common practice in video generation, we utilize a frozen pre-retired VAE~\cite{xing2024dynamicrafter} to encode RGB images, depth maps, and semantic maps into a unified latent space.
Given three input modalities: RGB images $\mathbf{I}_{i} \in \mathbb{R}^{H \times W \times 3}$, single-channel depth maps $\mathbf{D}_{i} \in \mathbb{R}^{H \times W \times 1}$, and K-channel semantic maps $\mathbf{S}_{i} \in \mathbb{R}^{H \times W \times K}$, to satisfy the VAE's input specifications, we convert the depth maps into pseudo-RGB images through channel replication, while transforming the semantic maps into RGB-compatible representations via colorization.
The encoding process can be described as:
\begin{equation}
\boldsymbol{z}_{\mathbf{I}} = \mathcal{E}({\mathbf{I}}_i), \quad \boldsymbol{z}_{\mathbf{D}} = \mathcal{E}({\mathbf{D}}_i^{\text{3ch}}), \quad \boldsymbol{z}_{\mathbf{S}} = \mathcal{E}({\mathbf{S}}_i^{\text{colorized}}),
\end{equation}
where $\mathcal{E}$ denotes the VAE encoder, ${\mathbf{D}}_i^{\text{3ch}}$ is the 3-channel depth map, and ${\mathbf{S}}_i^{\text{colorized}}$ is the colorized semantic map. 

During decoding, the VAE decoder reconstructs RGB, depth maps, and semantic maps from the latent space. For the depth map, we average the three channels of the decoder output to obtain the final single-channel depth map. For the semantic map, we revert the colorized output to the original labels via nearest-neighbor color matching. The decoding process can be described as follows:
\vspace{-0.3cm}
\begin{equation}
\begin{split}
{\mathbf{I}}_{i}^{\text{recon}} = \mathcal{D}(\boldsymbol{z}_{\mathbf{I}}), \quad {\mathbf{D}}_i^{\text{recon}} = \frac{1}{3} \sum_{c=1}^3 \mathcal{D}(\boldsymbol{z}_{\mathbf{D}})_c, \\
{\mathbf{S}}_i^{\text{recon}} = \arg\min_{k} \| \mathcal{D}(\boldsymbol{z}_{\mathbf{S}}) - \text{Color}(k) \|_2,
\end{split}
\end{equation}
\vspace{-0.1cm}
where $\mathcal{D}$ denotes the VAE decoder, and $\text{Color}(k)$ represents the predefined color corresponding to the $k$-th semantic class.\looseness=-1

\subsubsection{Training Phase of MDM}
As illustrated in Fig.~\ref{fig:pipeline}(a), the training phase of MDM begins with random sampling of a known camera trajectory $\{\mathbf{C}_{i}\}_{i=0}^N$ paired with its dense observations $\mathcal{I}=\{\mathbf{I}_{i}\}^{N}_{i=0}$.
Each view's geometric and semantic information is represented through corresponding depth maps $\mathcal{D}$ and semantic maps $\mathcal{S}$, denoted as $\mathcal{D}=\{\mathbf{D}_{i}\}^{N}_{i=0}$ and $\mathcal{S}=\{\mathbf{S}_{i}\}^{N}_{i=0}$, respectively. 
After fusion of LiDAR point clouds, we project point clouds into the image space using calibrated camera parameters, generating two sparse representations: sparse color images $\mathcal{P}_{\text{color}}=\{\mathbf{P}^{\text{color}}_{i}\}^{N}_{i=0}$ and sparse depth maps $\mathcal{P}_{\text{depth}}=\{\mathbf{P}^{\text{depth}}_{i}\}^{N}_{i=0}$.
To establish spatial consistency, we designate the first frame as the reference view and substitute its corresponding sparse representation, forming our sparse conditional input as follows:
\begin{equation}
\mathcal{P}_{\text{color}}=\{\mathbf{I}_{0}, \mathbf{P}^{{\text{color}}}_{1},\dots,\mathbf{P}^{{\text{color}}}_{N}\}.
\end{equation}

We particularly emphasize that preserving the latent features of reference images in the conditional input queue is crucial for maintaining consistency between the generated results and the original scene. This design becomes essential as sparse LiDAR points inherently lack sufficient information to constrain the upper regions of the images, which necessitates the incorporation of the dense reference image to ensure control throughout the generation process.

Then, the dense images, depth maps, and semantic maps are encoded into latent representation with the VAE encoder
$\boldsymbol{z}_{\mathbf{I}}=\mathcal{E}(\mathcal{I})$, $\boldsymbol{z}_{\mathbf{D}}=\mathcal{E}(\mathcal{D})$, and $\boldsymbol{z}_{\mathbf{S}}=\mathcal{E}(\mathcal{S})$. 

At each timestep $t$ in training phase, we add noise to the sampled data $\boldsymbol{z} \in \{\boldsymbol{z}_\mathbf{I},\boldsymbol{z}_\mathbf{D},\boldsymbol{z}_\mathbf{S}\}$, resulting in the noisy latent representations $\boldsymbol{z} \in \{\boldsymbol{z}_{\mathbf{I},t},\boldsymbol{z}_{\mathbf{D},t},\boldsymbol{z}_{\mathbf{S},t}\}$. 
Additionally, we encode the sparse color and depth inputs to obtain control signal: $\boldsymbol{y}_{\mathbf{I}} =\mathcal{E}(\mathcal{P^{\text{color}}})$ and $\boldsymbol{y}_{\mathbf{D}} =\mathcal{E}(\mathcal{P^{\text{depth}}})$. 
We then concatenate the noisy latent representations of each modality with the control signal as the input to the denoising diffusion network.
The model is trained using a v-prediction objective, where the target $\mathbf{v}_t$ is defined as:
\vspace{-0.1cm}
\begin{equation}
\mathbf{v}_{t} = \alpha_{t}\boldsymbol{\epsilon}_{t} -\sigma_t \boldsymbol{x}_{t},
\end{equation}
where $\boldsymbol{\epsilon}_{t} \sim \mathcal{N}(\mathbf{0}, \mathbf{I})$ denotes the sampled gaussian noise, $\alpha_t$ and $\sigma_t$ represent the time-dependent noise scheduling coefficients, and $\boldsymbol{x}_t$ corresponds to the noisy input modality requiring denoising. 
The training objective is defined as:
\vspace{-0.1cm}
\begin{equation}
\mathcal{L}=\mathbb{E}_{\boldsymbol{x},\boldsymbol{\epsilon},t,\boldsymbol{s}}\left\|\mathcal{F}_{\theta}\left(\boldsymbol{x}_{t}, \boldsymbol{x}_{\text{ref}}, \boldsymbol{y}_{\mathbf{I}}, \boldsymbol{y}_{\mathbf{D}}, \boldsymbol{s}\right)-\mathbf{v}_{t}\right\|_{2}^{2},
\end{equation}
$\boldsymbol{y}_{\mathbf{I}}$ and $\boldsymbol{y}_{\mathbf{D}}$ denote the sparse conditional inputs that guide the generation process, while $\boldsymbol{x}_{\text{ref}}$ provides the reference image for appearance consistency. The vector $\boldsymbol{s}$ represents the input modality that is denoised, with $\boldsymbol{x}_t$ indicating its noisy state in the diffusion time step $t$.

\begin{figure}[t]
\centering
\includegraphics[width=\linewidth]{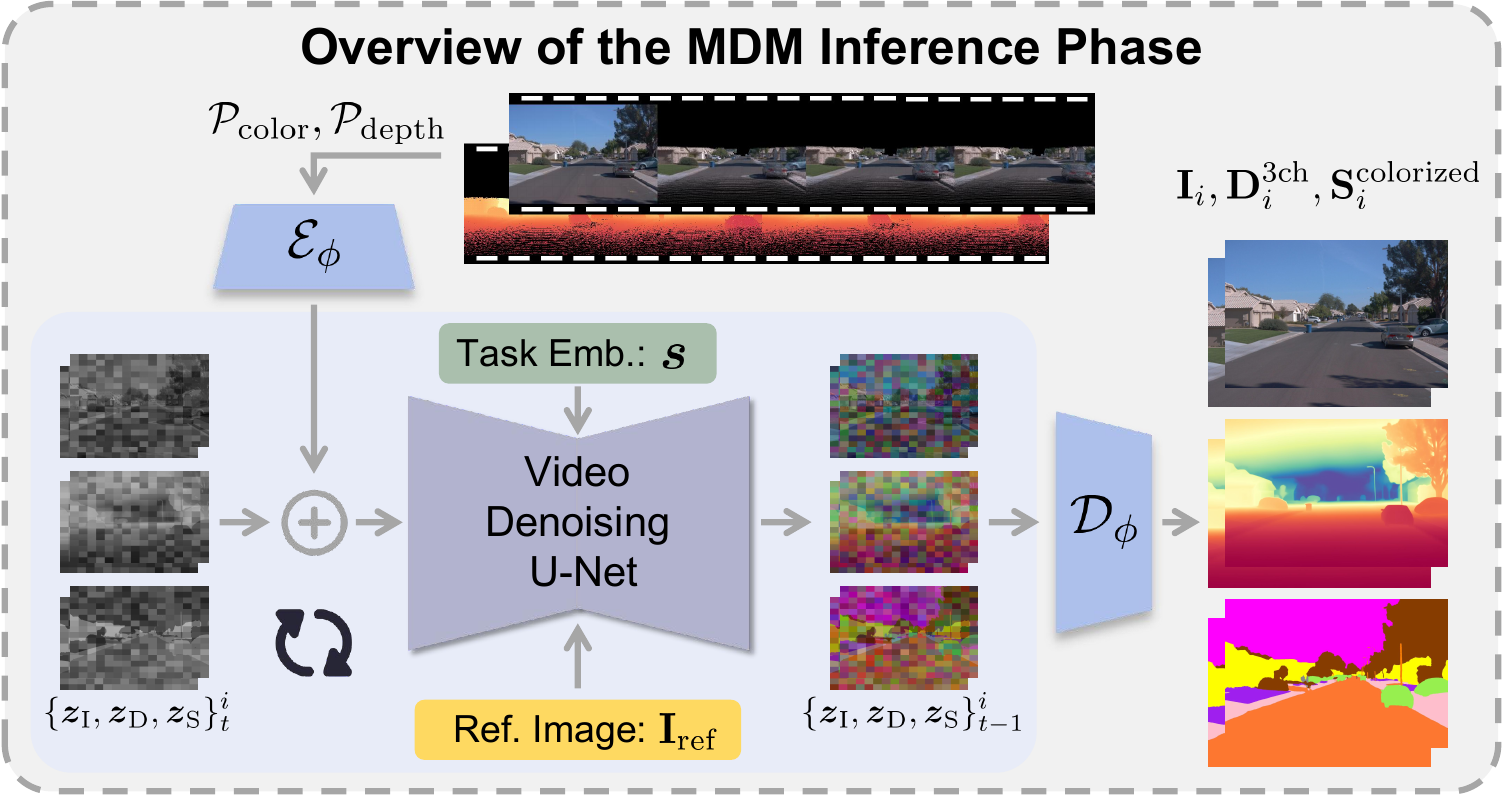}
\caption{\textbf{Illustration of inference phase.} During inference, our multi-modal diffusion model takes a dense reference image and a sequence of condition images as inputs, generating the corresponding sequences of RGB, depth, and semantic maps.
}
\label{fig:infer_v2}
\vspace{-0.5cm}
\end{figure}

\subsection{Inferring Phase of MDM}
After MDM training, it can be directly applied as a generalized model to new scenarios without requiring per-scene optimization.
Specifically, we first project the fused point clouds to novel viewpoints, yielding sparse color $\{\mathbf{I}_{\text{sparse},i}\}_{i=1}^{N}$ and depth maps $\{\mathbf{D}_{\text{sparse},i}\}_{i=0}^{N}$. We subsequently select adjacent observed images as reference views to form the input conditional as:
\vspace{-0.1cm}
\begin{equation}
\mathcal{I}_{\text{cond}} = \{ \mathbf{I}_{\text{ref}}\} + \{ \mathbf{I}_{\text{sparse},i}\}_{i=1}^{N}, \quad \mathcal{D}_{\text{cond}} = \{\mathbf{D}_{\text{sparse},i}\}_{i=0}^{N}.
\end{equation}

As shown in Fig.~\ref{fig:infer_v2}, we initialize the generation process by sampling standard Gaussian noise $z_{T}$ from a normal distribution. This noise vector is subsequently combined with encoded conditional inputs through channel-wise concatenation, where modality-specific guidance signals $s$ are injected to steer different generation modalities.
Consistent with the diffusion schedule used during training, we perform iterative denoising steps on the fused latent representation through our learned conditional diffusion process. The refined latent is finally decoded by the pre-trained VAE decoder to dense predictions in the pixel space, including color images, depth maps, and semantic maps.

\subsection{Urban Scene View Synthesis}\label{subsec:gs}
Current urban scene view synthesis methods tend to overfit training viewpoints, leading to significant artifacts when synthesizing novel views with significant perspective shifts (\eg, lane changes). 
This limitation arises from insufficient viewpoint diversity in training data, which severely constrains the generalization capability of 3DGS frameworks.
To address this, we propose to leverage our multi-modal diffusion model to generate novel view RGB images, depth maps, and semantic maps for providing rich supervisory for GS training and thereby improving the quality of novel view synthesis.\looseness=-1

Specifically, we first generate novel camera trajectories, which is followed by generating corresponding novel view RGB images, depth maps, and semantic maps.
To better utilize virtual viewpoints, during each training iteration, we select virtual viewpoints with a probability of $\theta$. The overall optimization objective is defined as:
\vspace{-0.1cm}
\begin{equation}
\mathcal{L}_{\mathrm{input}} = \lambda_1 \mathcal{L}_1 + \lambda_{\mathrm{ssim}} \mathcal{L}_{\mathrm{ssim}} + \lambda_{\mathrm{depth}} \mathcal{L}_{\mathrm{depth}},
\end{equation}
\begin{equation}
\mathcal{L}_{\mathrm{virtual}} = \lambda_{\mathrm{v_{color}}} \mathcal{L}_{\mathrm{lpips}} + \lambda_{\mathrm{v_{depth}}} \mathcal{L}_{\mathrm{depth}} + \lambda_{\mathrm{v_{sem}}} \mathcal{L}_{\mathrm{sem}},
\end{equation}
where $\mathcal{L}_1$, $\mathcal{L}_{\mathrm{ssim}}$, $\mathcal{L}_{\mathrm{lpips}}$, and $\mathcal{L}_{\mathrm{depth}}$ denote the L1, SSIM, LPIPS and depth loss, respectively. For virtual viewpoints, $\mathcal{L}_{\mathrm{lpips}}$, $\mathcal{L}_{\mathrm{depth}}$, and $\mathcal{L}_{\mathrm{sem}}$ represent the LPIPS loss, depth loss, and Cross-Entropy loss, respectively.

\section{Experiments}

\subsection{Implementation Details}
Our multi-modal diffusion model (MDM) builds upon~\cite{xing2024dynamicrafter}, following the original training configuration with v-objective.
Specifically, we initialize MDM with the pre-trained weights of~\cite{xing2024dynamicrafter} while adapting its 8-channel input layers to accommodate our 12-channel inputs through weights replication and scaling.
The training phase comprises two sequential stages: 
First, we train MDM with a resolution of $320\times512$ using a batch size of 8, learning rate of 1.0e-5, and Adam optimizer for 20,000 iterations on 8$\times$ NVIDIA H20 GPUs, incorporating 20\% random reference image dropout and a DDPM noise scheduler with 1,000 diffusion steps.
Subsequently, we freeze the temporal layers and fine-tune the spatial components at a resolution of $576\times1024$ with reduced batch size 4 (same learning rate) for 8,000 iterations. 
For inference, we employ the DDIM~\cite{song2020denoising} scheduler with 50-step sampling to balance efficiency and quality. 
The 3D Gaussian Splatting (3DGS) module adapts from~\cite{yan2024streetgs}, training each scene for 30,000 iterations on L20 GPUs with a 20\% virtual viewpoint sampling probability to enhance geometric generalization.

\subsection{Datasets}
Our MDM is trained on the Open Waymo Dataset (WOD)~\cite{sun2020scalability} with 32 selected sequences. 
To generate sparse RGB and depth inputs, we aggregate lidars across the entire scene and project temporally fused point clouds onto each frame. 
For semantic map ground-truth, we employ the~\cite{xie2021segformer} to produce pseudo semantic maps for each frame. 
To obtain dense depth ground-truth, we first aggregate six consecutive LiDAR frames and project them into the current viewpoint, resulting in sparse LiDAR depth maps. 
Following established depth completion practices, these sparse maps are first densified using~\cite{liu2024depthlab}, then geometrically aligned with the original LiDAR measurements through least-squares optimization to ensure consistency.
\vspace{-0.1cm}

\begin{table}[!t]
\centering
\resizebox{\linewidth}{!}{
\begin{tabular}{l|cc|cc|cc}
\toprule
\multicolumn{1}{l|}{\multirow{2}{*}{}} & \multicolumn{2}{c|}{Shift 2m} & \multicolumn{2}{c|}{Shift 3m} & \multicolumn{2}{c}{Shift 4m} \\
\cmidrule(lr){2-3} \cmidrule(lr){4-5} \cmidrule(lr){6-7}
Method  & FID $\downarrow$ & FVD $\downarrow$ &  FID $\downarrow$ & FVD $\downarrow$ & FID $\downarrow$ &  FVD $\downarrow$ \\
\midrule
S$^3$Gaussian~\cite{huang2024S3Gaussian}   & 169.08  &  2031.97 &  191.38  & 2543.47 & 206.99 & 2663.56 \\
3DGS$^*$~\cite{gaussian_splatting}   & 101.63  & 1062.57 & 117.36  & 1379.68 & 135.10 & 1675.16  \\
EmerNeRF~\cite{yang2023emernerf} & 73.58 & 645.90 &  93.09 & 983.94 & 110.94 & 1250.56 \\
StreetGS~\cite{yan2024streetgs} & 69.79    & \underline{563.00}    & 87.71     & 898.05    & 100.05    & 1269.99 \\
FreeVS~\cite{wang2024freevs} & 92.08    & 652.09 &  95.91 & 784.74  & 117.68   & \underline{801.20}  \\
\midrule
\rowcolor{gray!10}Ours-R & \textbf{50.02}     & 671.24    & \textbf{49.42} & \underline{777.78}    & \textbf{56.52}     & \textbf{770.39} \\
\rowcolor{gray!10}Ours-S & \underline{54.14}  & \textbf{509.34}    & \underline{58.85}     & \textbf{659.27}    & \underline{65.44} & 808.85  \\
\bottomrule
\end{tabular}
}
\vspace{-5pt}
\caption{
\textbf{Quantitative comparisons on novel view synthesis.}
\textbf{Ours-R} denotes the outputs generated by our multi-modal diffusion model, while \textbf{Ours-S} integrates the results rendered from optimized 3D Gaussian Splatting Representation. 
3DGS$^*$ denotes the results, excluding failure cases.
Our method demonstrates superior quality even under large viewpoint displacements, such as ±4-meter lateral shifts.
Furthermore, \textbf{Ours-R} exhibits exceptional robustness to variations in camera pose.
(\textbf{Bold} figures indicate the best and \underline{underlined} figures indicate the second best) 
}
\vspace{-10pt}
\label{tab:all_results}
\end{table}

\subsection{Experimental Results}
\subsubsection{Multi-model Results}
\begin{figure*}[!t]
\setlength\tabcolsep{0.5 pt}
\centering
\scalebox{0.8}{
\begin{tabular}{cccccc}

Ref. Image & Color Cond. & Depth Cond. & Color & Depth & Semantic \\

\begin{tabular}{l}\includegraphics[width=0.198\linewidth]{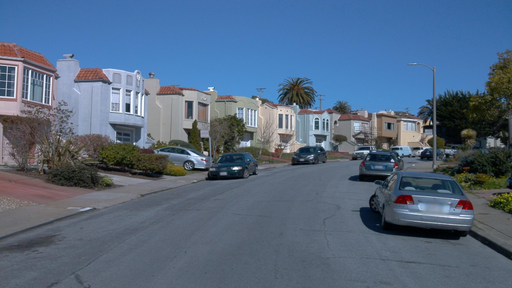}\end{tabular} &
\begin{tabular}{l}\includegraphics[width=0.198\linewidth]{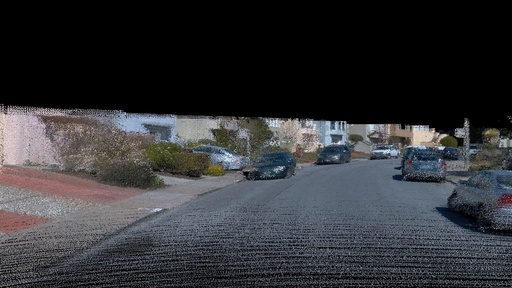}\end{tabular} &
\begin{tabular}{l}\includegraphics[width=0.198\linewidth]{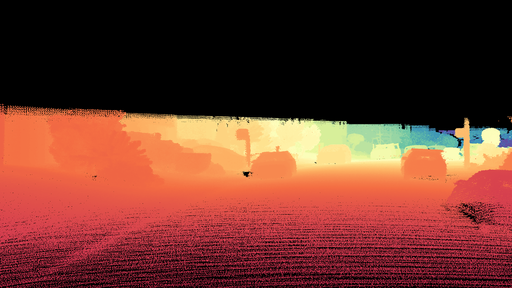}\end{tabular} &
\begin{tabular}{l}\includegraphics[width=0.198\linewidth]{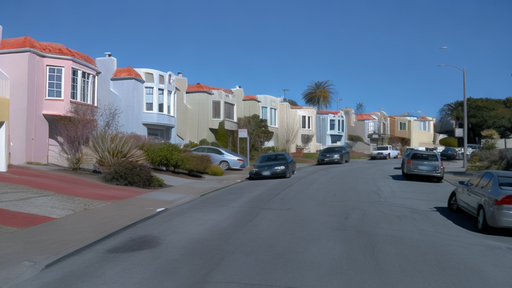}\end{tabular} &
\begin{tabular}{l}\includegraphics[width=0.198\linewidth]{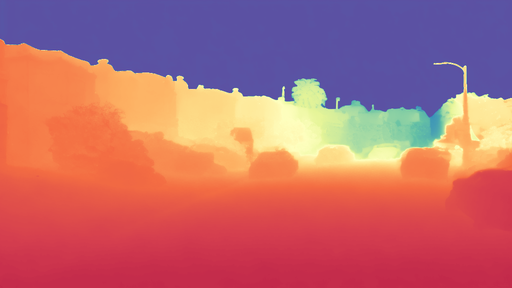}\end{tabular} &
\begin{tabular}{l}\includegraphics[width=0.198\linewidth]{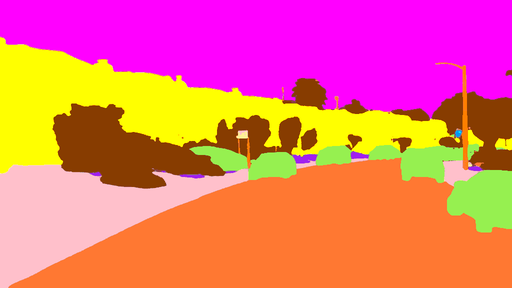}\end{tabular} \\

\begin{tabular}{l}\includegraphics[width=0.198\linewidth]{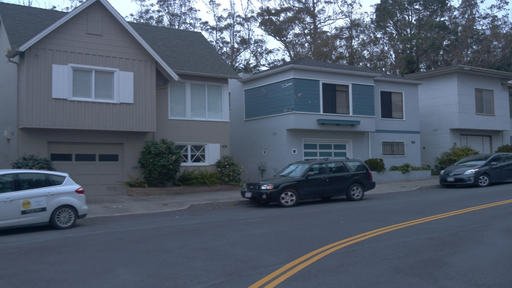}\end{tabular} &
\begin{tabular}{l}\includegraphics[width=0.198\linewidth]{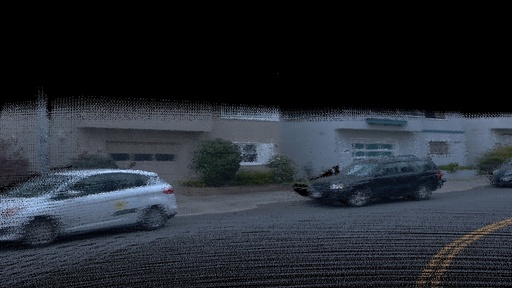}\end{tabular} &
\begin{tabular}{l}\includegraphics[width=0.198\linewidth]{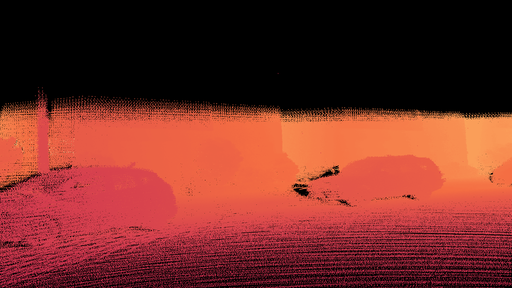}\end{tabular} &
\begin{tabular}{l}\includegraphics[width=0.198\linewidth]{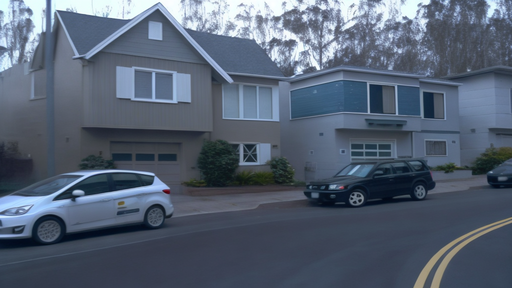}\end{tabular} &
\begin{tabular}{l}\includegraphics[width=0.198\linewidth]{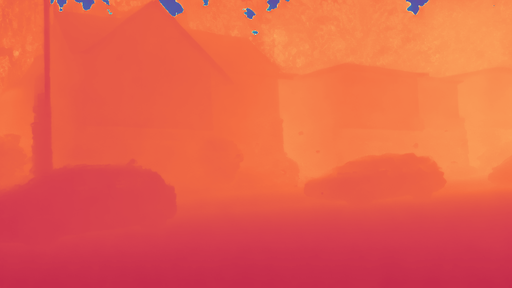}\end{tabular} &
\begin{tabular}{l}\includegraphics[width=0.198\linewidth]{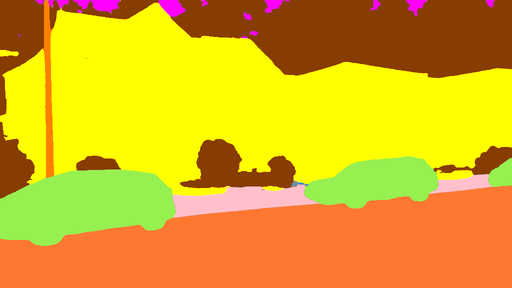}\end{tabular} \\


\begin{tabular}{l}\includegraphics[width=0.198\linewidth]{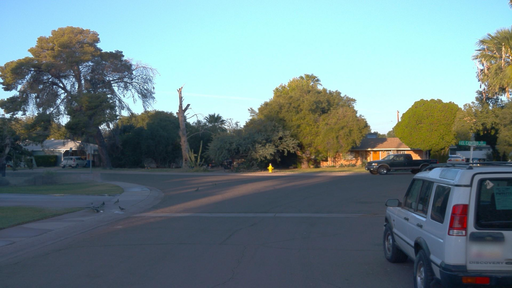}\end{tabular} &
\begin{tabular}{l}\includegraphics[width=0.198\linewidth]{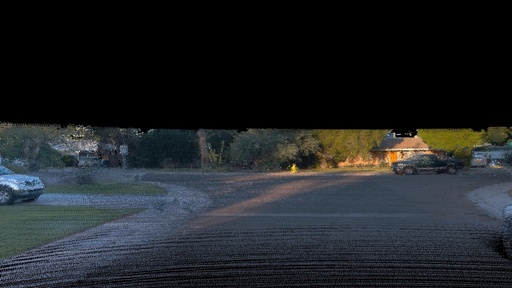}\end{tabular} &
\begin{tabular}{l}\includegraphics[width=0.198\linewidth]{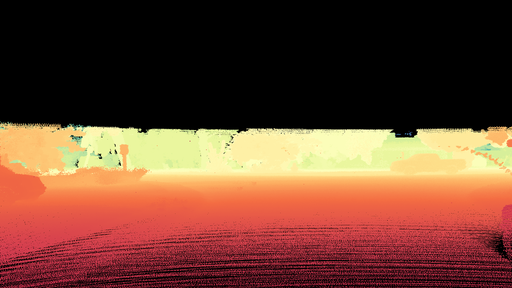}\end{tabular} &
\begin{tabular}{l}\includegraphics[width=0.198\linewidth]{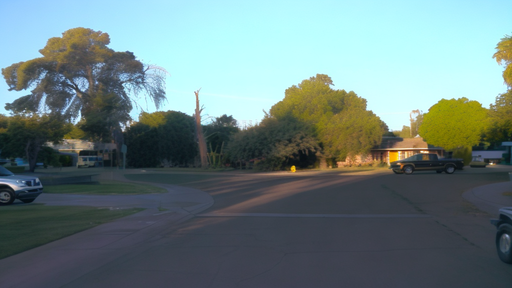}\end{tabular} &
\begin{tabular}{l}\includegraphics[width=0.198\linewidth]{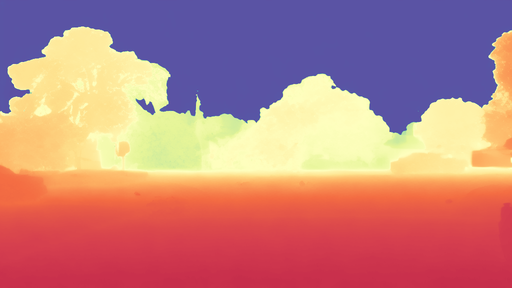}\end{tabular} &
\begin{tabular}{l}\includegraphics[width=0.198\linewidth]{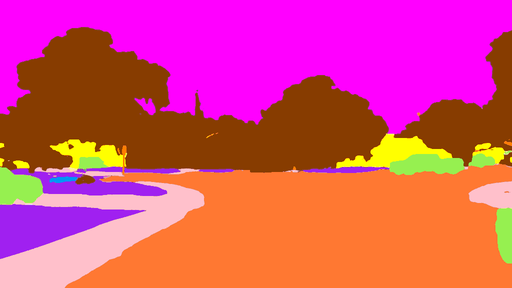}\end{tabular} \\

\begin{tabular}{l}\includegraphics[width=0.198\linewidth]{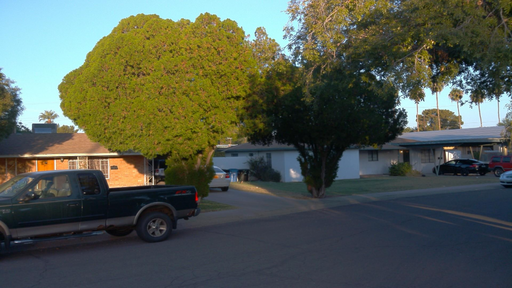}\end{tabular} &
\begin{tabular}{l}\includegraphics[width=0.198\linewidth]{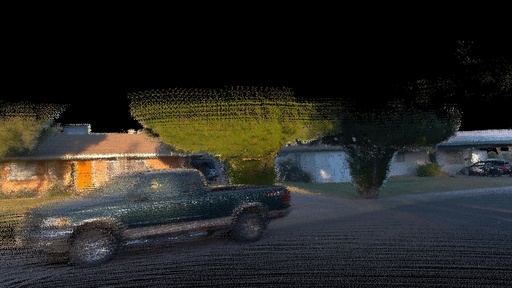}\end{tabular} &
\begin{tabular}{l}\includegraphics[width=0.198\linewidth]{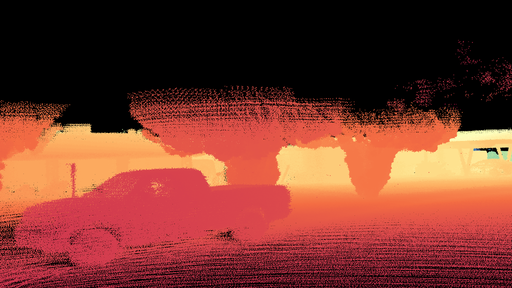}\end{tabular} &
\begin{tabular}{l}\includegraphics[width=0.198\linewidth]{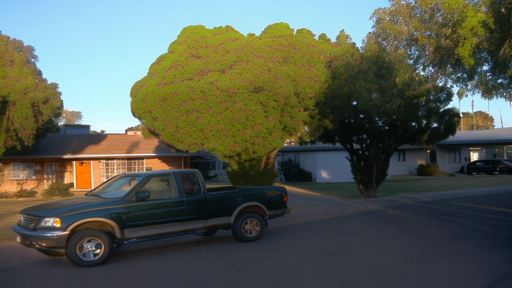}\end{tabular} &
\begin{tabular}{l}\includegraphics[width=0.198\linewidth]{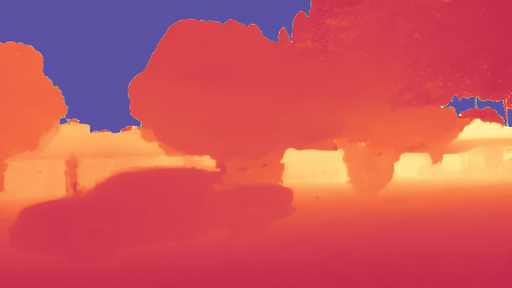}\end{tabular} &
\begin{tabular}{l}\includegraphics[width=0.198\linewidth]{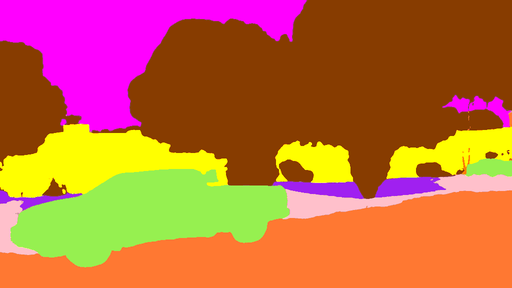}\end{tabular} \\



\end{tabular}}
\caption{\textbf{Visualization of multi-modal results.}
Given the reference imagea and sparse conditions, we present the visualized multi-modal novel-view synthesis results (color, depth, and semantic map). The depth maps are visualized in the range of [$0$, $100m$]. The controllable and photorealistic results highlight the robust multi-modal synthesis capabilities of our approach, even under extreme viewpoint variations.
}
\label{fig:vis_modal_results}
\end{figure*}

This paper introduces a novel framework for multi-modal novel view synthesis in street scenes under extreme viewpoint changes. 
As shown in Fig.~\ref{fig:vis_modal_results}, our MDM enables feed-forward novel view synthesis from sparse depth and color inputs while generating consistent RGB, depth, and semantic maps.
This view-aligned multi-modal generation capability directly supports autonomous driving applications through synthetic data generation and enhanced 3D reconstruction.\looseness=-1

\begin{figure}[!t]
\centering
\includegraphics[width=\linewidth]{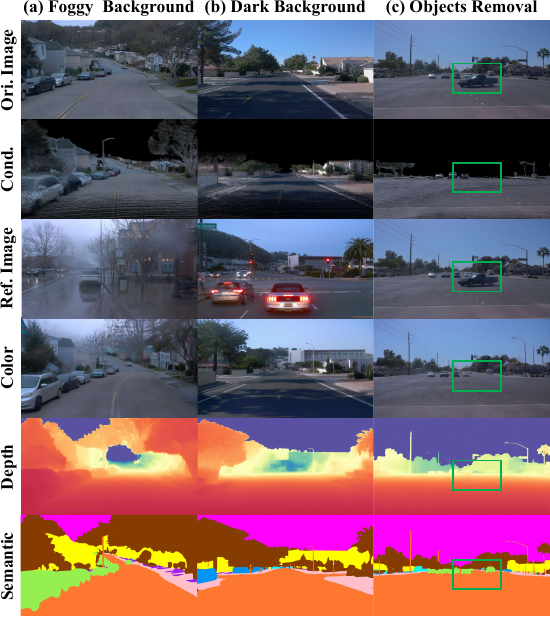}
\caption{
\textbf{Visualization of editing results.}
The figure demonstrates the ability of MDM to edit the background, such as transforming it into a foggy or dark setting, and to remove objects. 
MDM can effectively modify the target area based on the reference image while maintaining consistency with original images.
}
\label{fig:edit}
\vspace{-0.4cm}
\end{figure}

As shown in Fig.~\ref{fig:edit}, our model's controllability allows precise editing, including background replacement, lane changes, and object removal. It also generates aligned multi-modal outputs (RGB, depth, semantics) that maintain consistency during editing.


\subsubsection{Benchmark Evaluation}
In this section, we compare our approach with state-of-the-art street view synthesis methods, as quantitatively presented in Tab.~\ref{tab:all_results}. Here, Ours-R represents the outputs of MDM, while Ours-S denotes the MuDG approach, which integrates MDM with the 3DGS module.
Since ground truth is unavailable for novel view synthesis outputs in the lane-shifting setting, we evaluate rendering quality using Fréchet Inception Distance (FID)~\cite{heusel2017gans} and Fréchet Video Distance (FVD)~\cite{wang2018video}. As shown in Tab.~\ref{tab:all_results}, our method outperforms existing methods by a large margin and achieves state-of-the-art performance.
Notably, Ours-R demonstrates greater stability under extreme viewpoint shifts (-4m lateral) than Ours-S. This is due to Ours-S’s susceptibility to artifacts at scene boundaries, whereas Ours-R maintains photorealistic consistency even in these challenging regions.
Here, we observe that the 3DGS module fails in two dynamic scenes due to the rare initialization of Gaussian points and the metrics for 3DGS$^*$ are calculated based on the remaining scenes.

\begin{figure*}[!t]
\setlength\tabcolsep{0.5 pt}
\centering
\scalebox{0.8}{
\begin{tabular}{cccccc}

Input view & 3DGS & S3GS & Street Gaussians & Ours-R & Ours-S \\
\begin{tabular}{l}\includegraphics[width=0.198\linewidth]{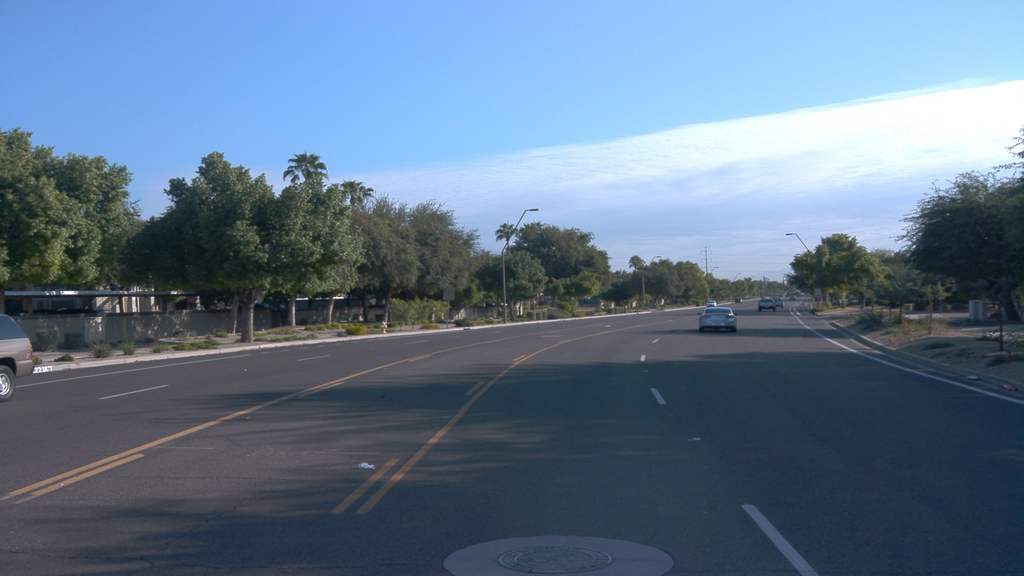}\end{tabular} &
\begin{tabular}{l}\includegraphics[width=0.198\linewidth]{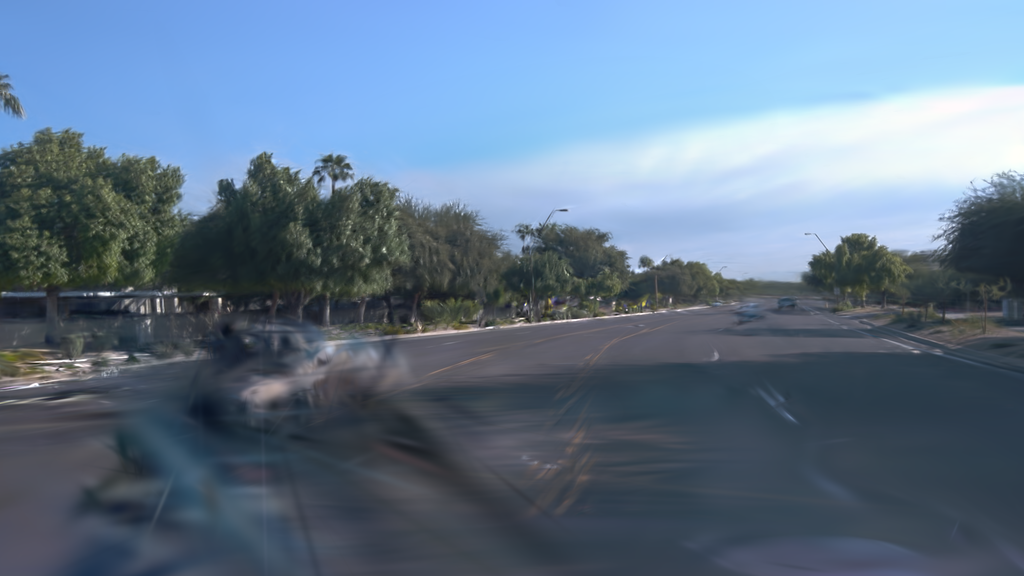}\end{tabular} &
\begin{tabular}{l}\includegraphics[width=0.198\linewidth]{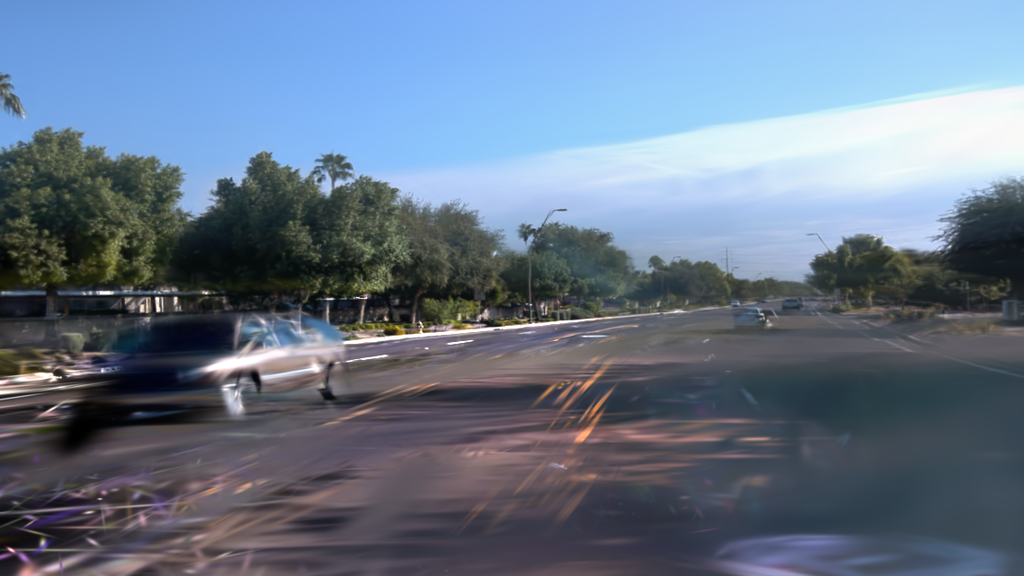}\end{tabular} &
\begin{tabular}{l}\includegraphics[width=0.198\linewidth]{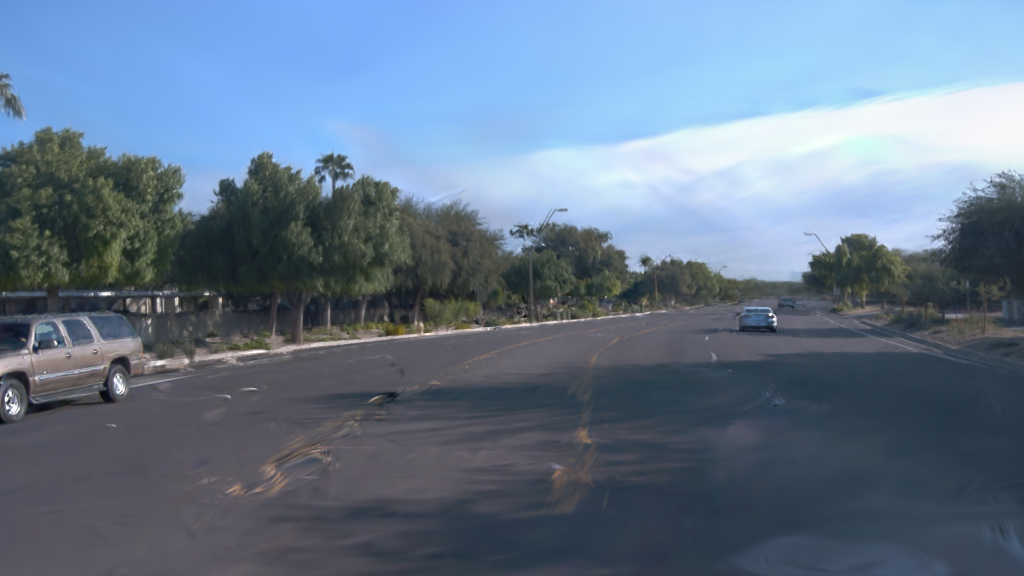}\end{tabular} &
\begin{tabular}{l}\includegraphics[width=0.198\linewidth]{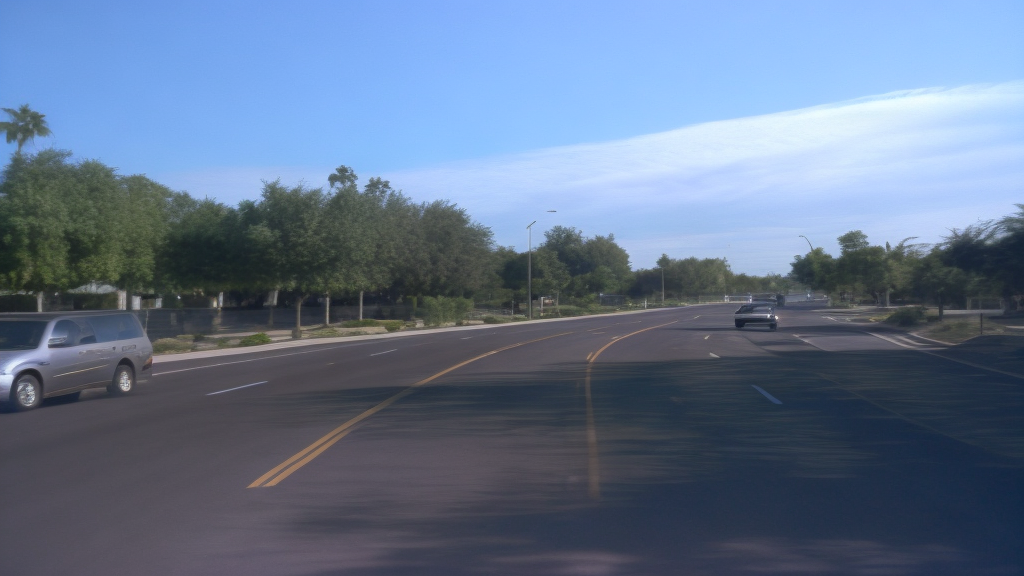}\end{tabular} &
\begin{tabular}{l}\includegraphics[width=0.198\linewidth]{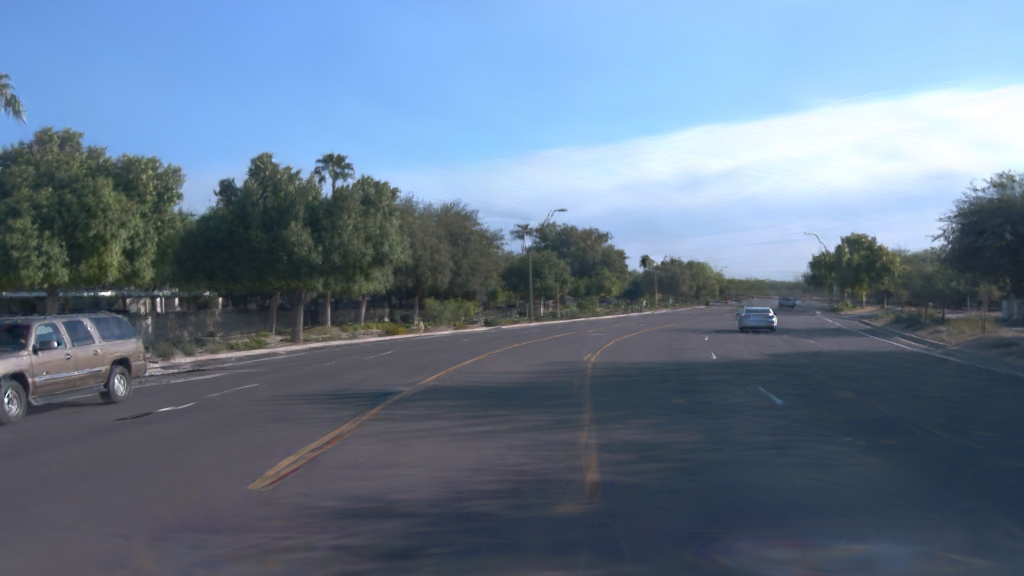}\end{tabular} \\

\begin{tabular}{l}\includegraphics[width=0.198\linewidth]{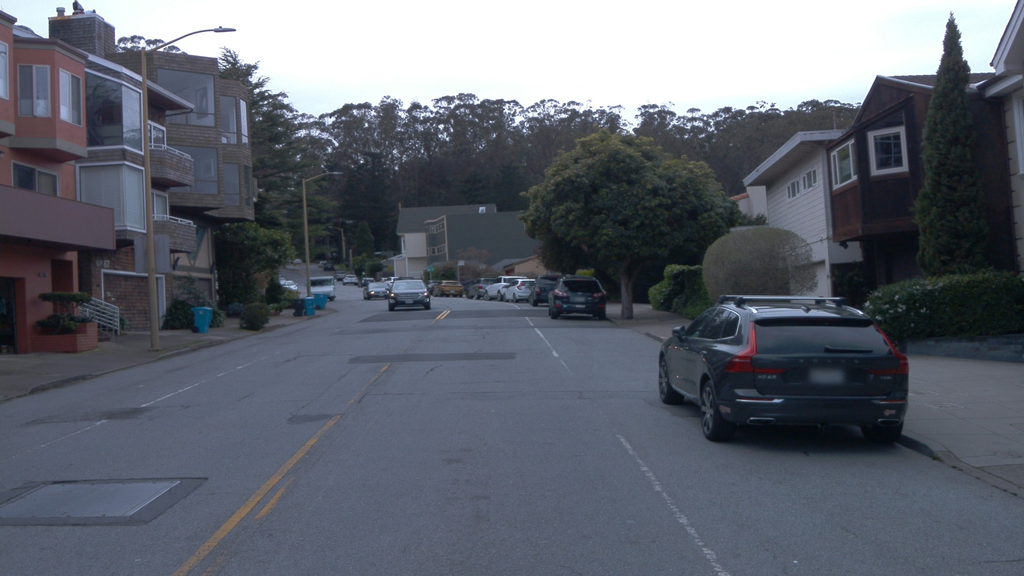}\end{tabular} &
\begin{tabular}{l}\includegraphics[width=0.198\linewidth]{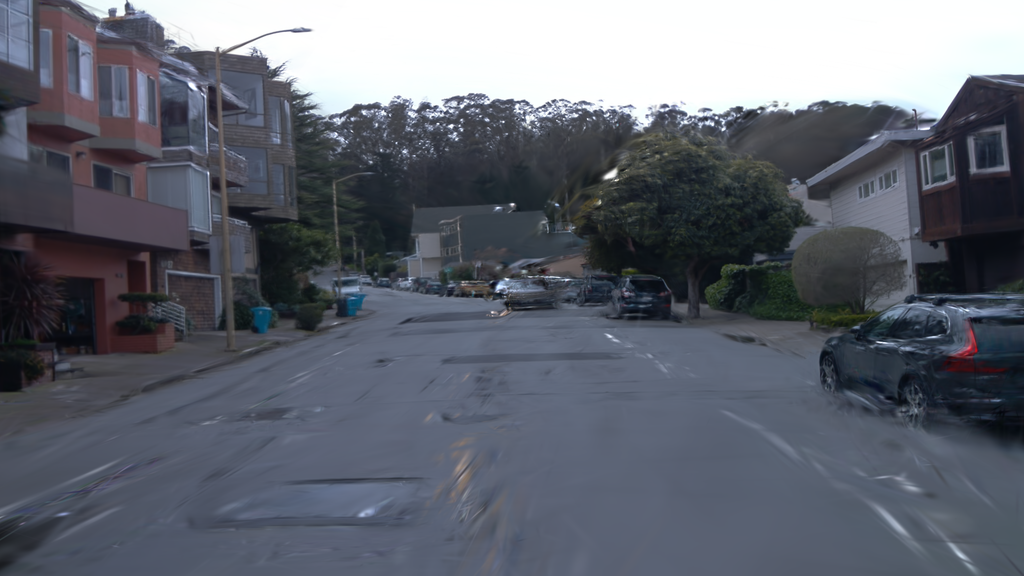}\end{tabular} &
\begin{tabular}{l}\includegraphics[width=0.198\linewidth]{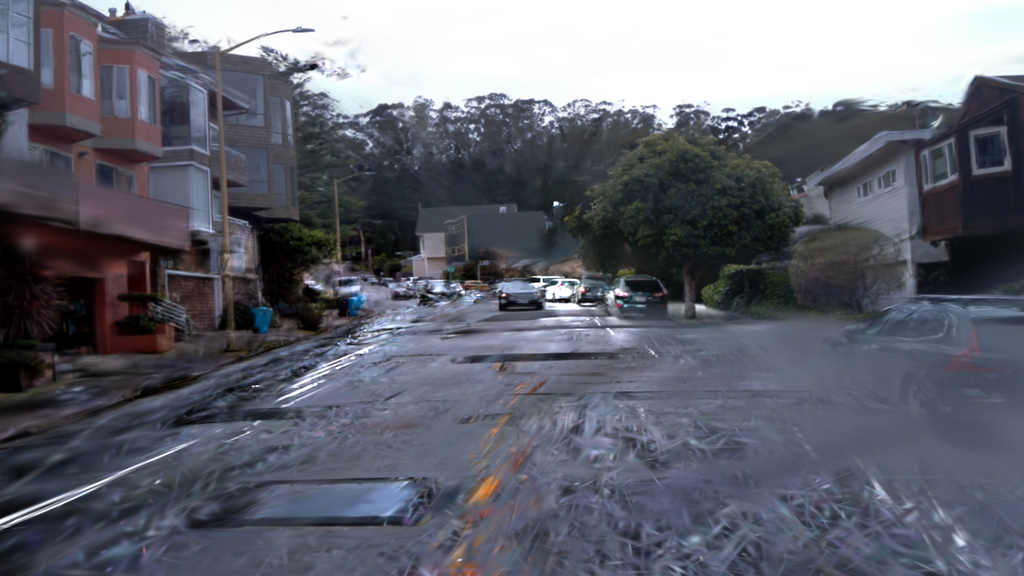}\end{tabular} &
\begin{tabular}{l}\includegraphics[width=0.198\linewidth]{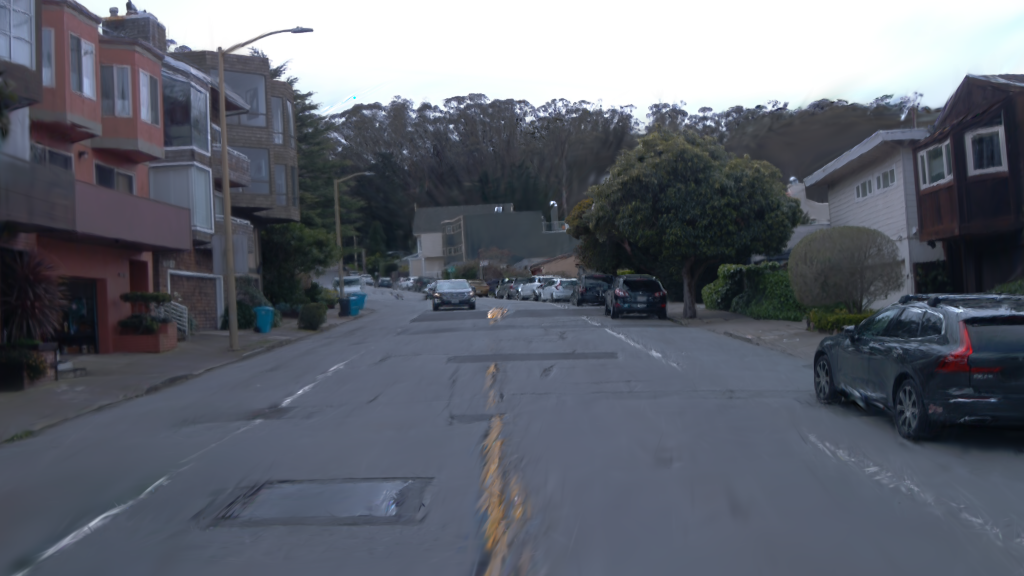}\end{tabular} &
\begin{tabular}{l}\includegraphics[width=0.198\linewidth]{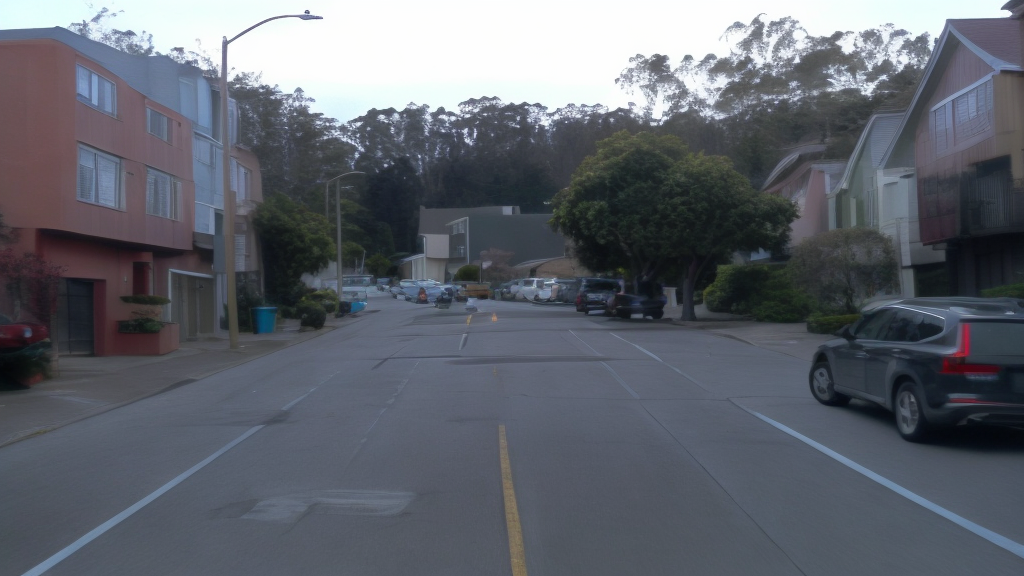}\end{tabular} &
\begin{tabular}{l}\includegraphics[width=0.198\linewidth]{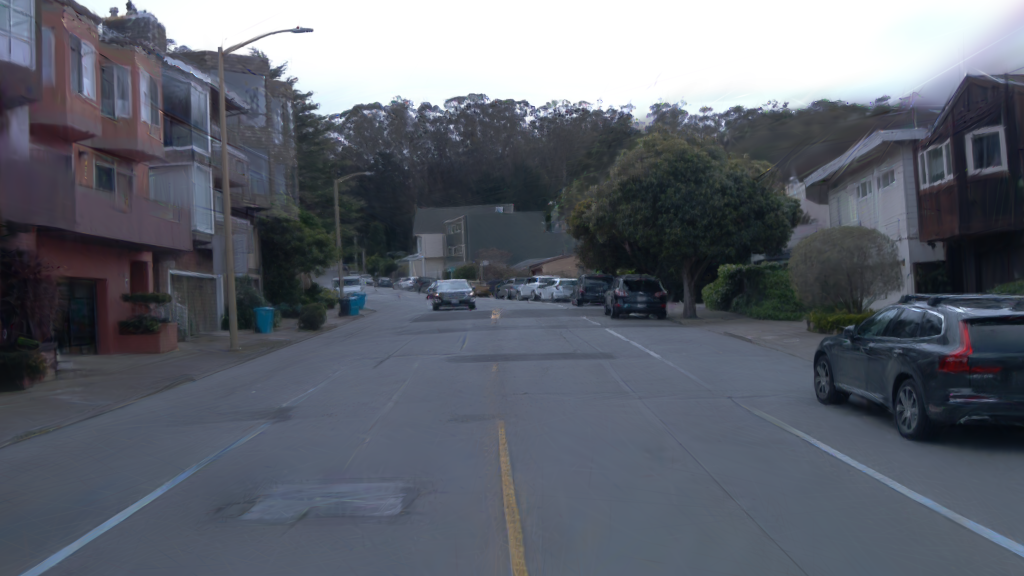}\end{tabular} \\

\begin{tabular}{l}\includegraphics[width=0.198\linewidth]{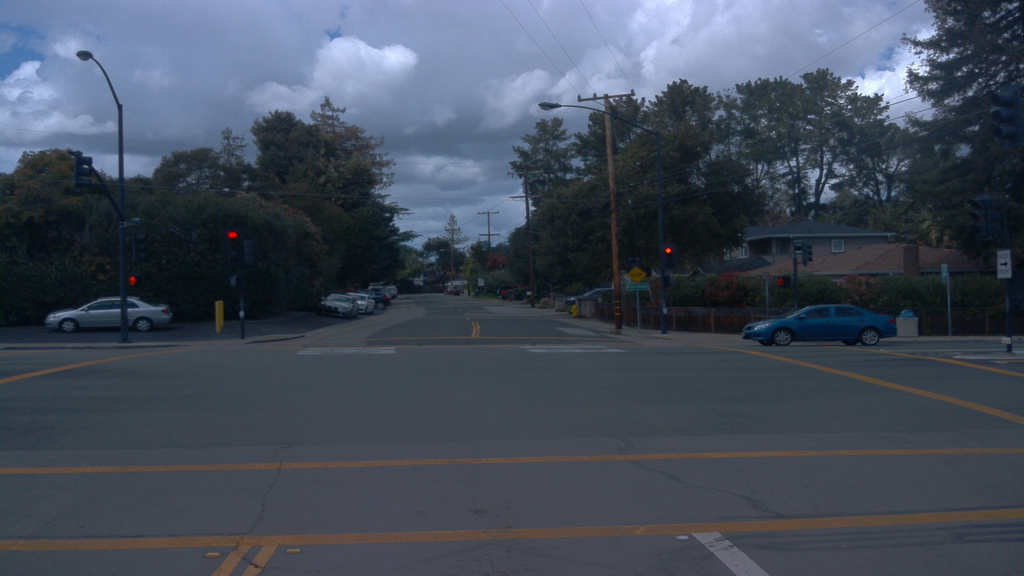}\end{tabular} &
\begin{tabular}{l}\includegraphics[width=0.198\linewidth]{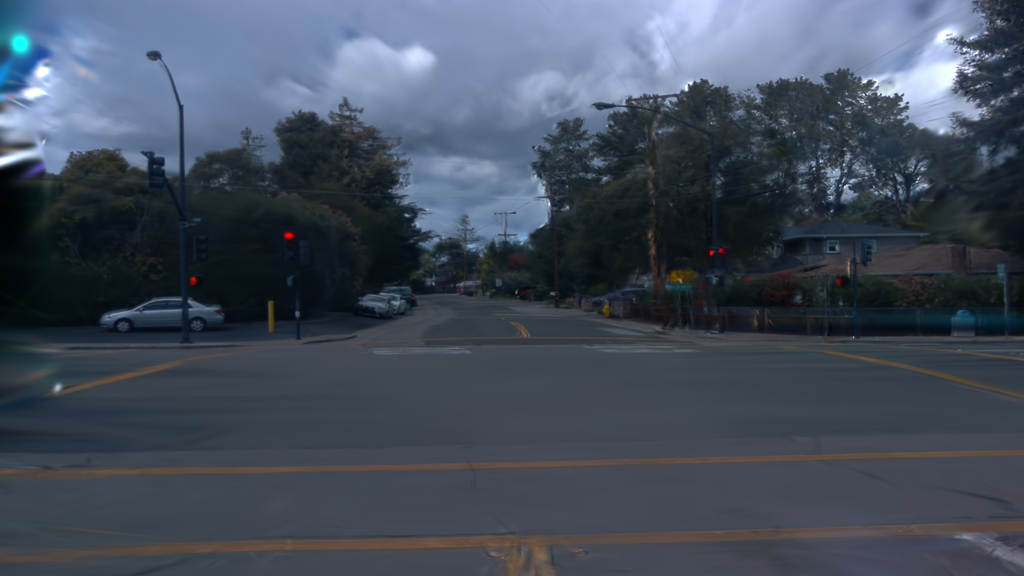}\end{tabular} &
\begin{tabular}{l}\includegraphics[width=0.198\linewidth]{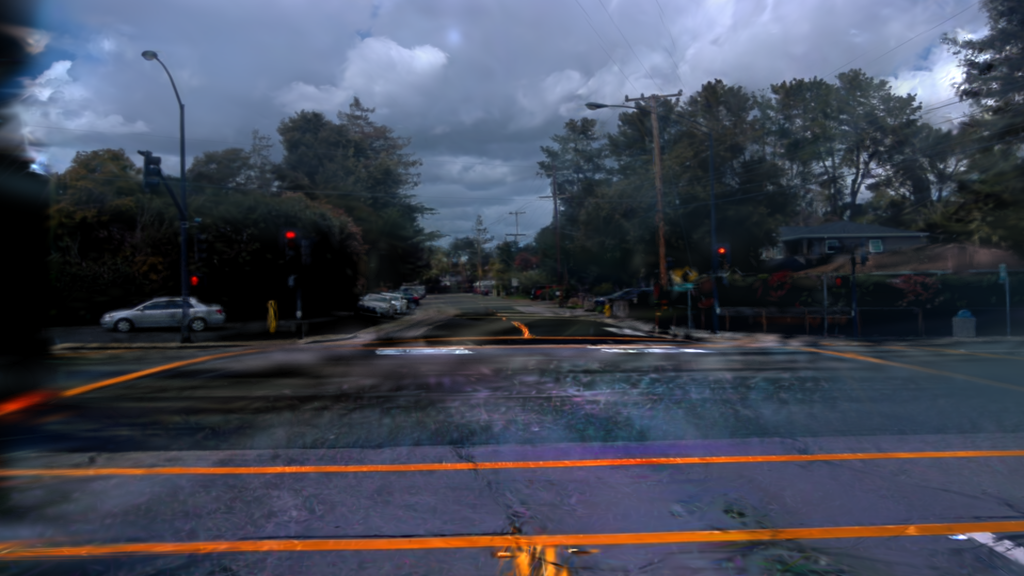}\end{tabular} &
\begin{tabular}{l}\includegraphics[width=0.198\linewidth]{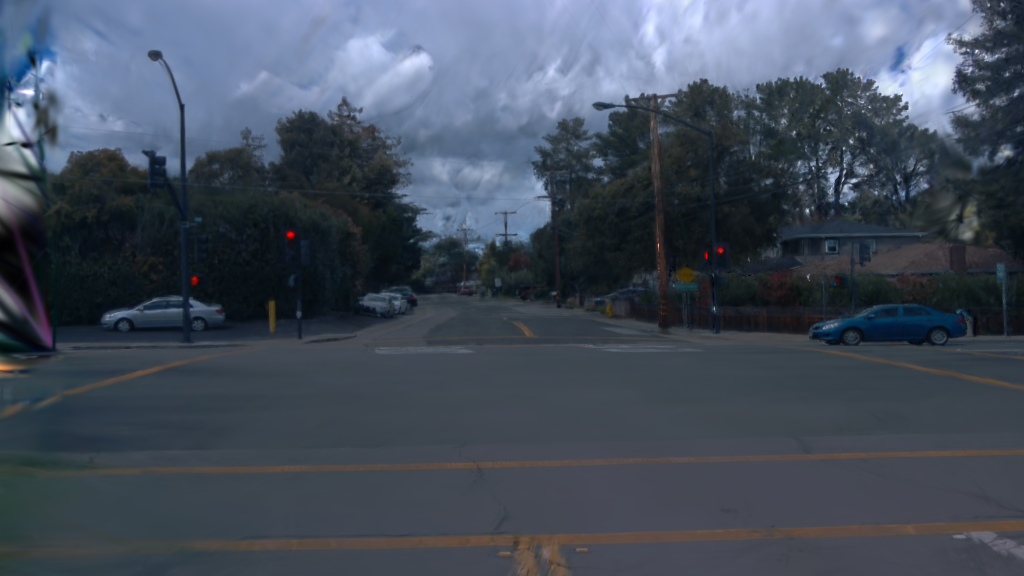}\end{tabular} &
\begin{tabular}{l}\includegraphics[width=0.198\linewidth]{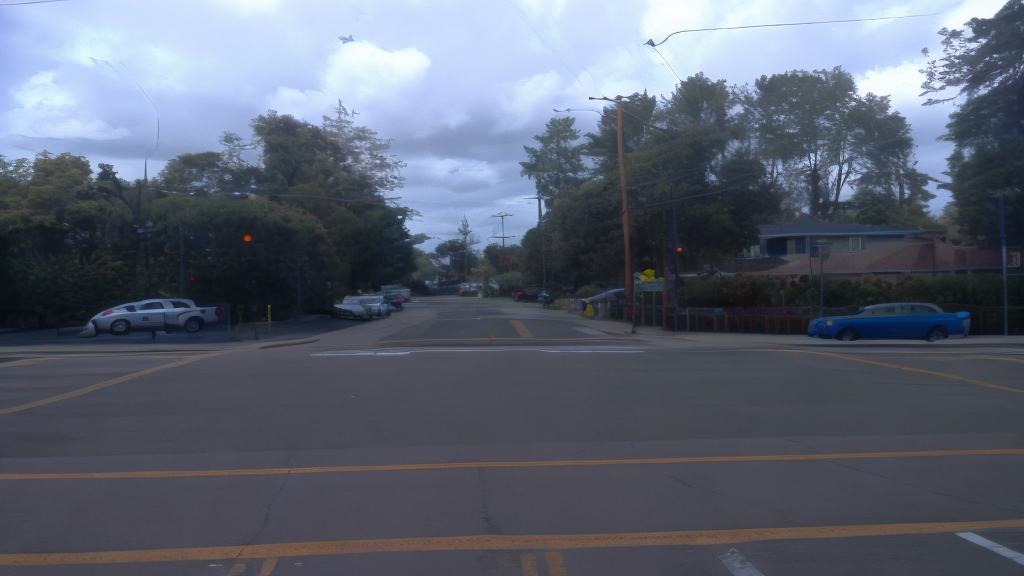}\end{tabular} &
\begin{tabular}{l}\includegraphics[width=0.198\linewidth]{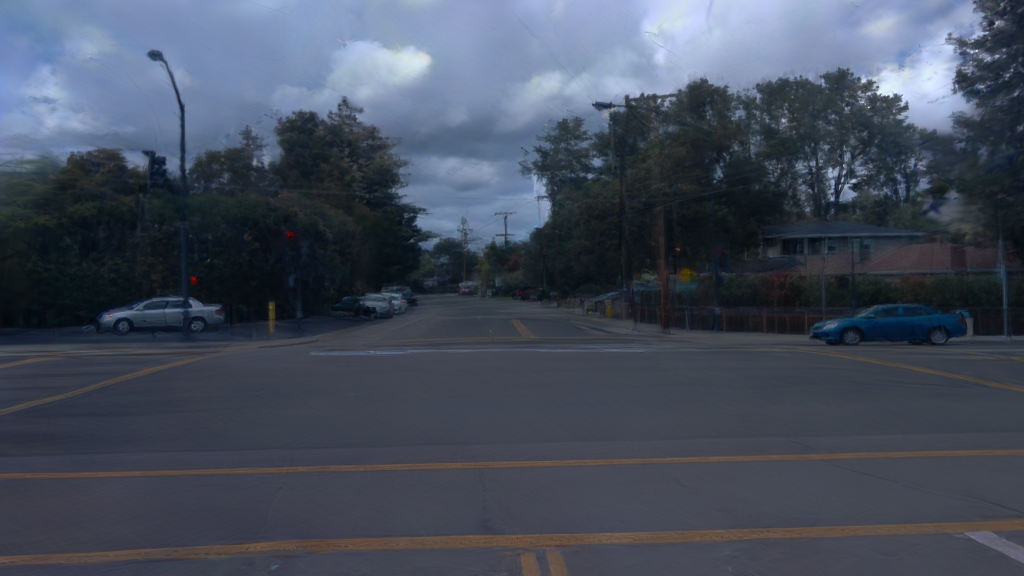}\end{tabular} \\

\begin{tabular}{l}\includegraphics[width=0.198\linewidth]{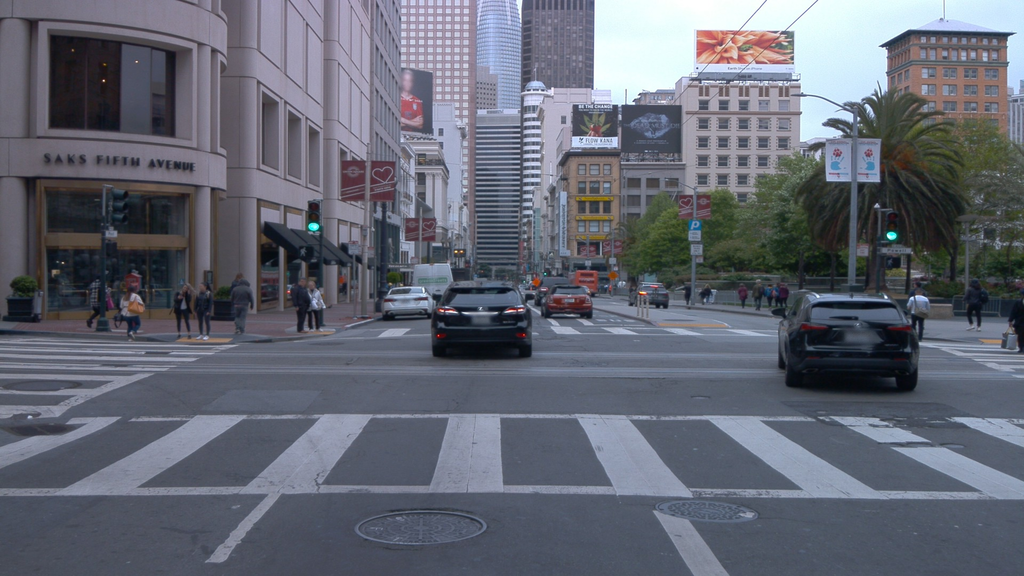}\end{tabular} &
\begin{tabular}{l}\includegraphics[width=0.198\linewidth]{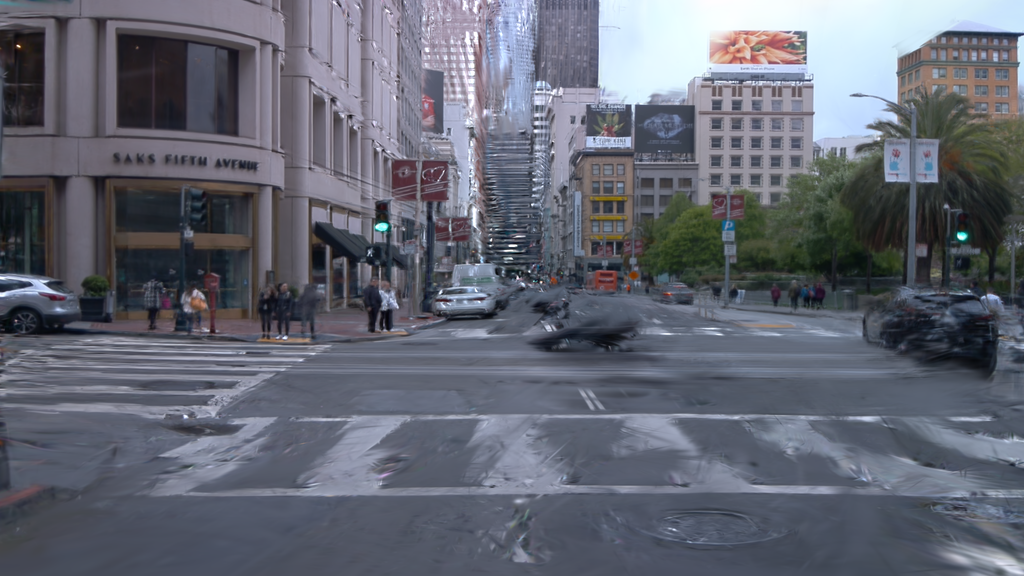}\end{tabular} &
\begin{tabular}{l}\includegraphics[width=0.198\linewidth]{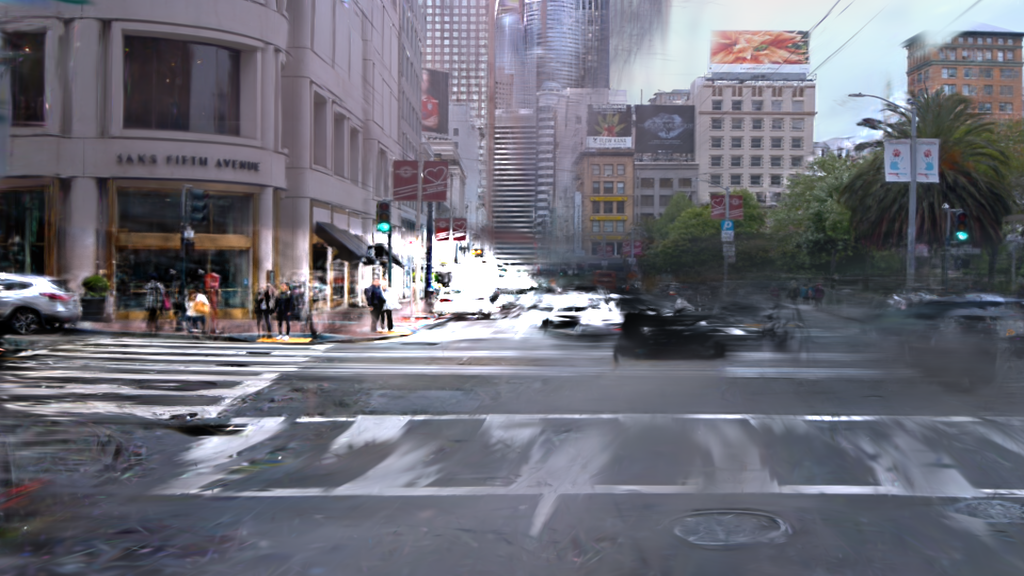}\end{tabular} &
\begin{tabular}{l}\includegraphics[width=0.198\linewidth]{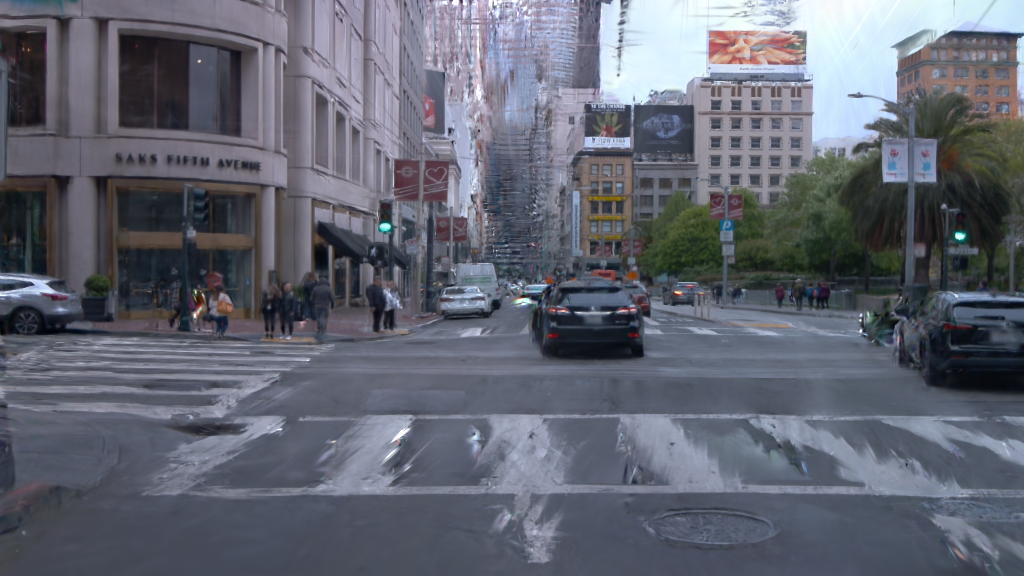}\end{tabular} &
\begin{tabular}{l}\includegraphics[width=0.198\linewidth]{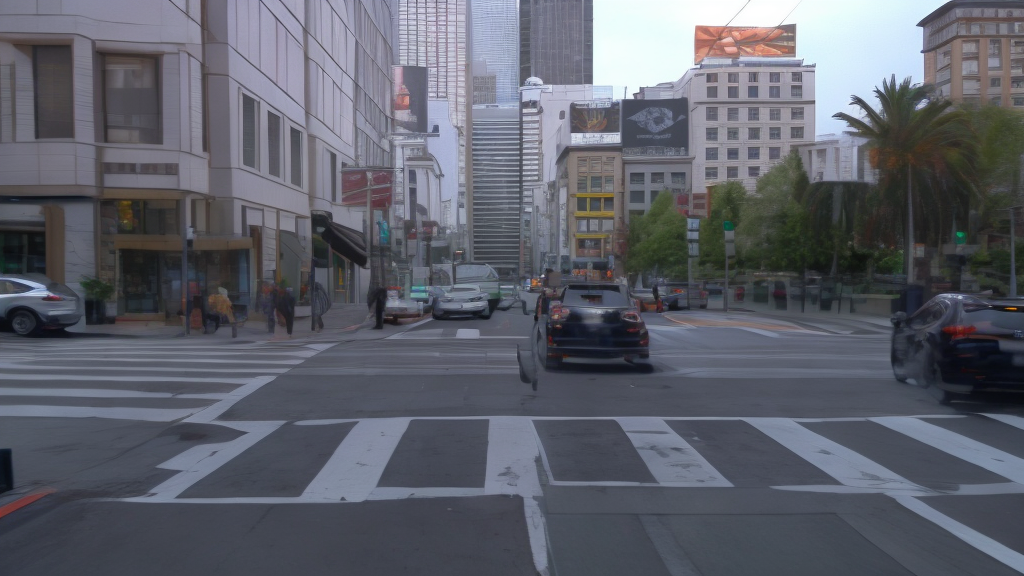}\end{tabular} &
\begin{tabular}{l}\includegraphics[width=0.198\linewidth]{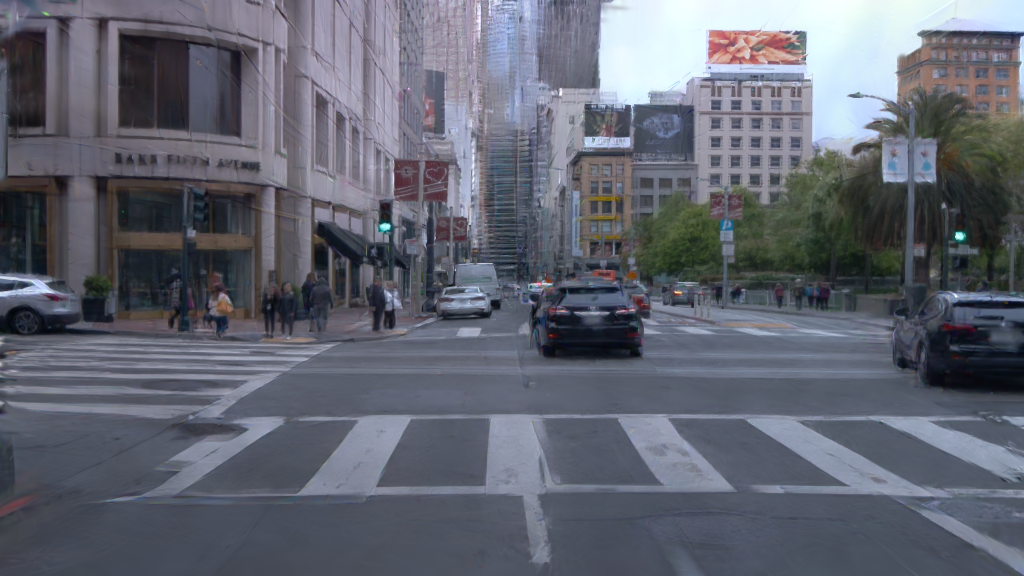}\end{tabular} \\

\begin{tabular}{l}\includegraphics[width=0.198\linewidth]{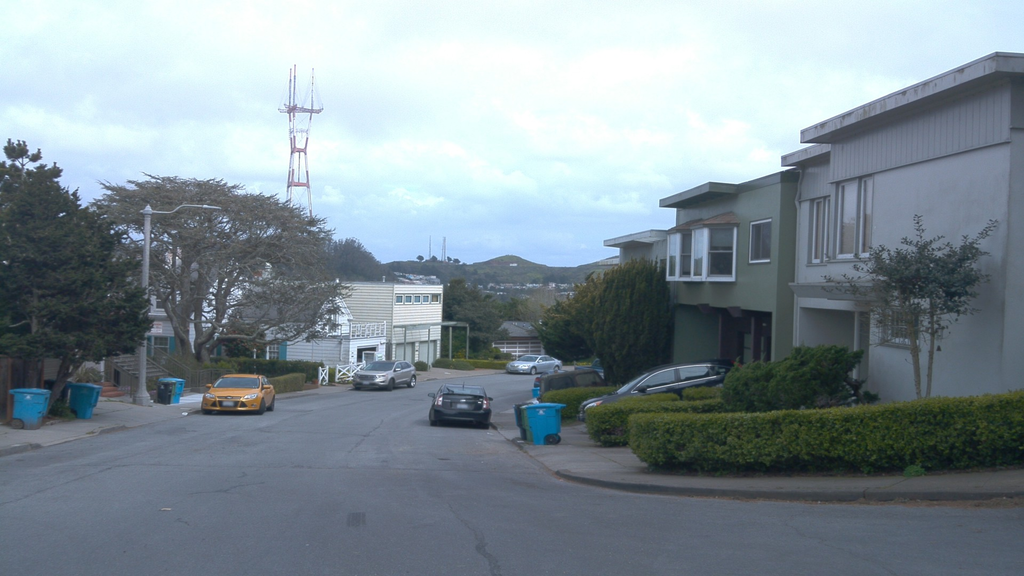}\end{tabular} &
\begin{tabular}{l}\includegraphics[width=0.198\linewidth]{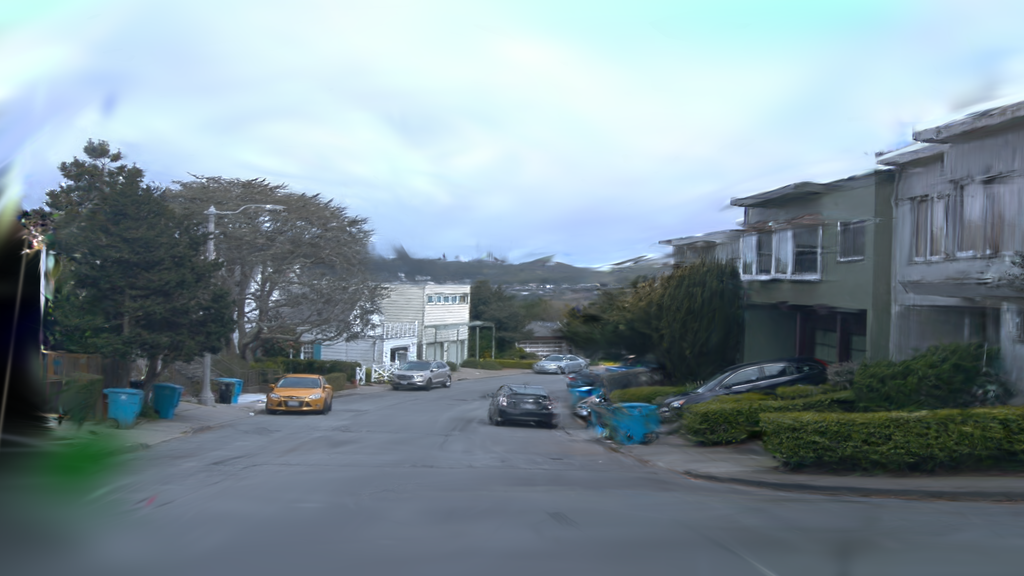}\end{tabular} &
\begin{tabular}{l}\includegraphics[width=0.198\linewidth]{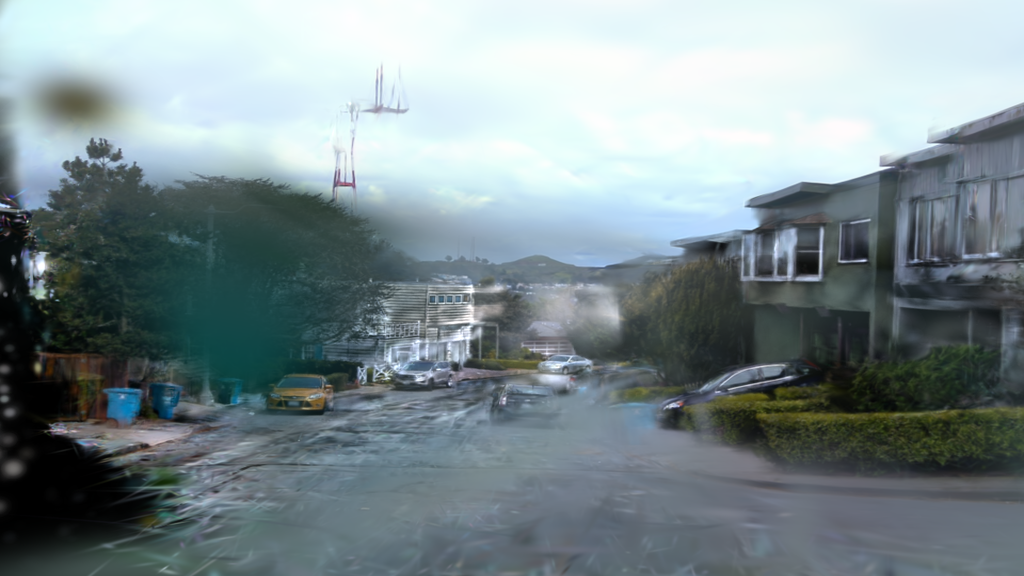}\end{tabular} &
\begin{tabular}{l}\includegraphics[width=0.198\linewidth]{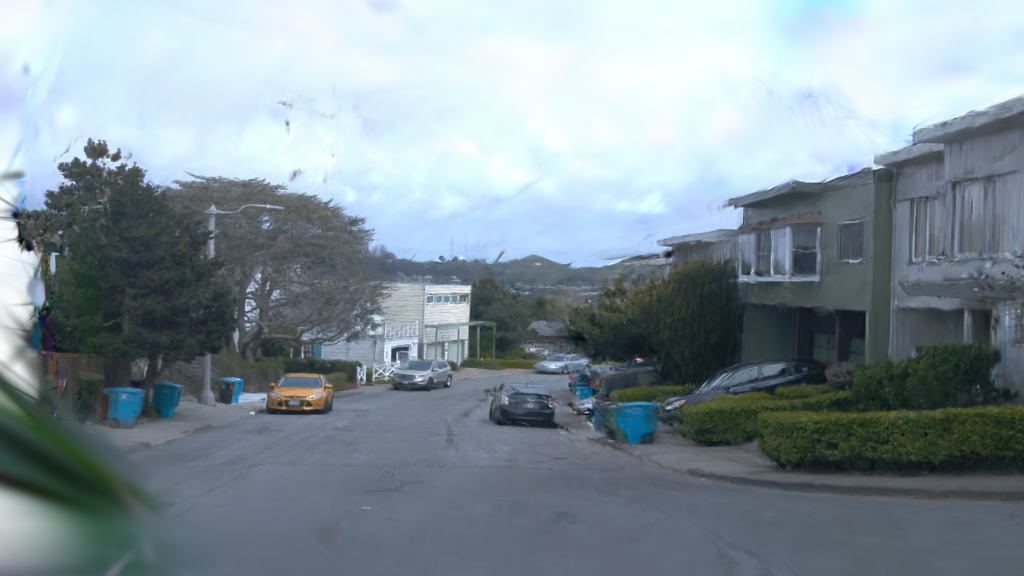}\end{tabular} &
\begin{tabular}{l}\includegraphics[width=0.198\linewidth]{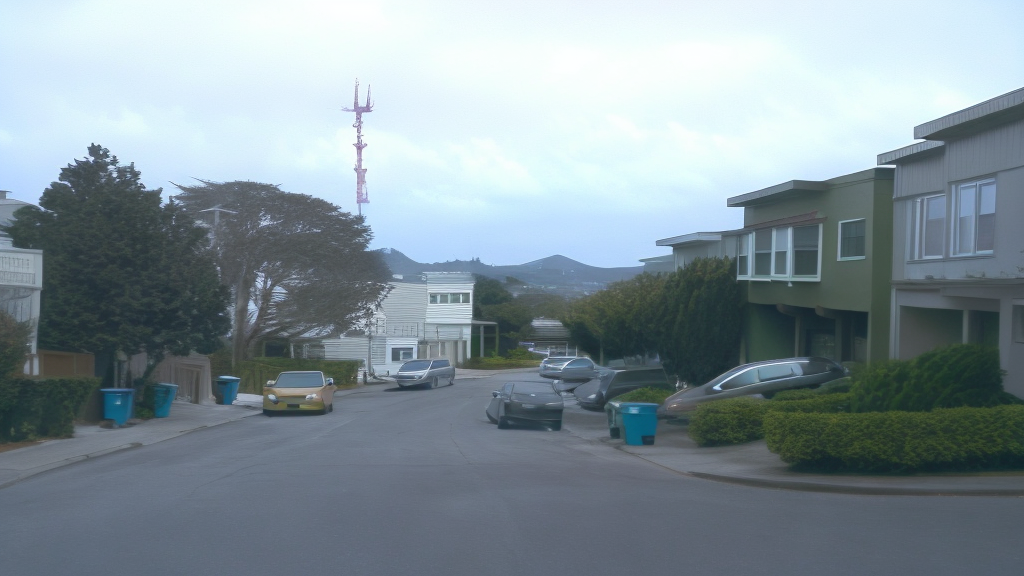}\end{tabular} &
\begin{tabular}{l}\includegraphics[width=0.198\linewidth]{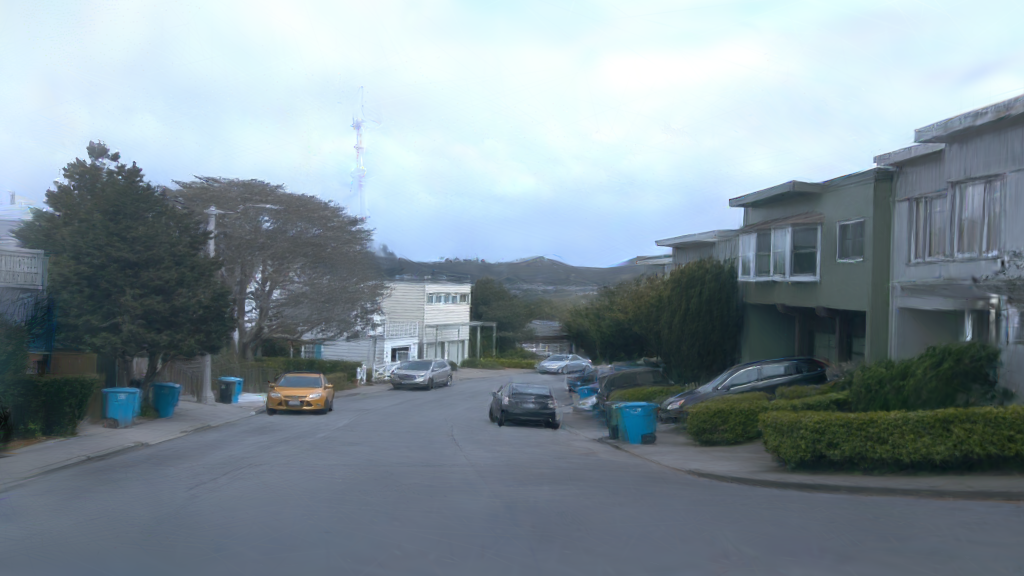}\end{tabular} \\

\begin{tabular}{l}\includegraphics[width=0.198\linewidth]{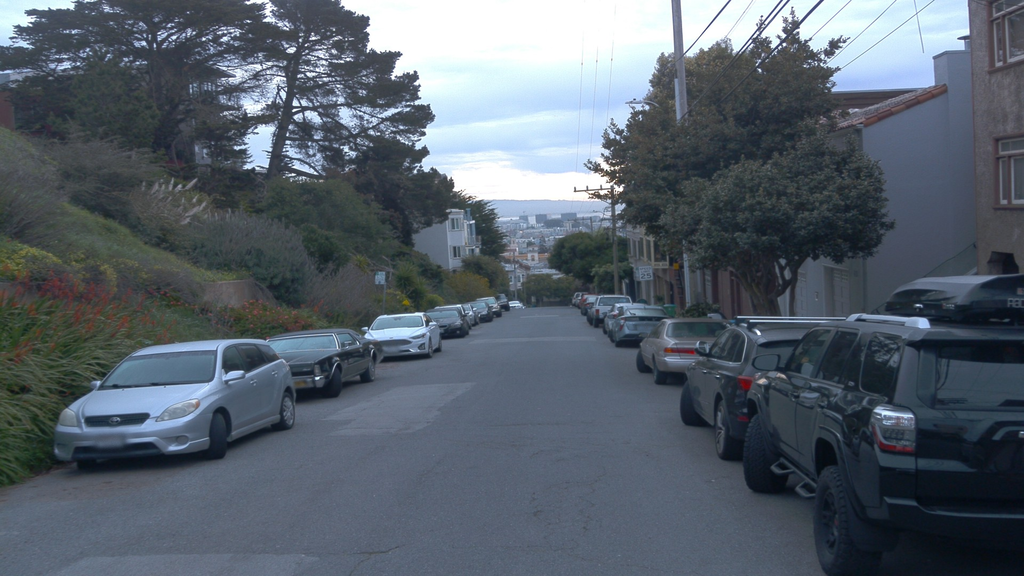}\end{tabular} &
\begin{tabular}{l}\includegraphics[width=0.198\linewidth]{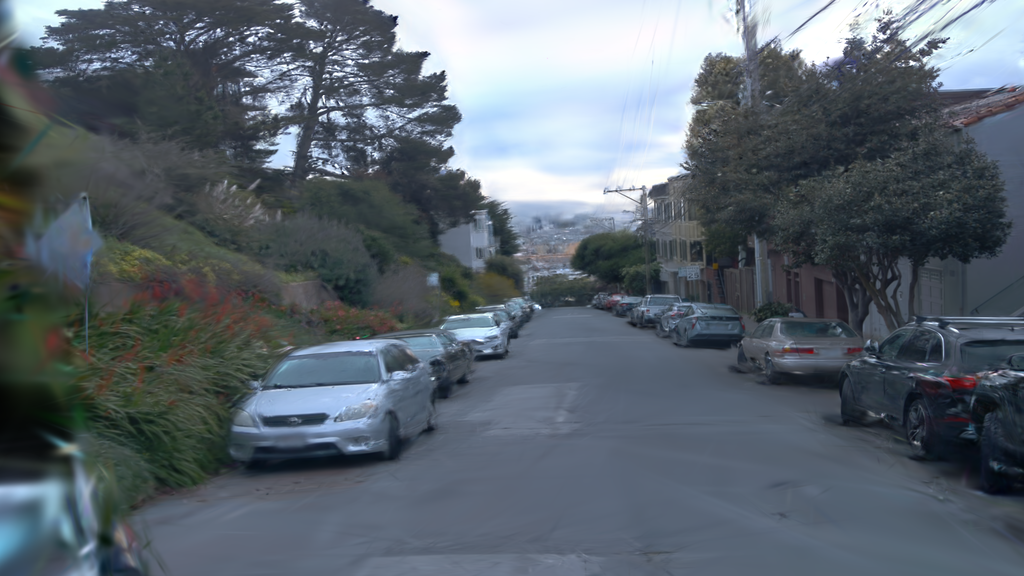}\end{tabular} &
\begin{tabular}{l}\includegraphics[width=0.198\linewidth]{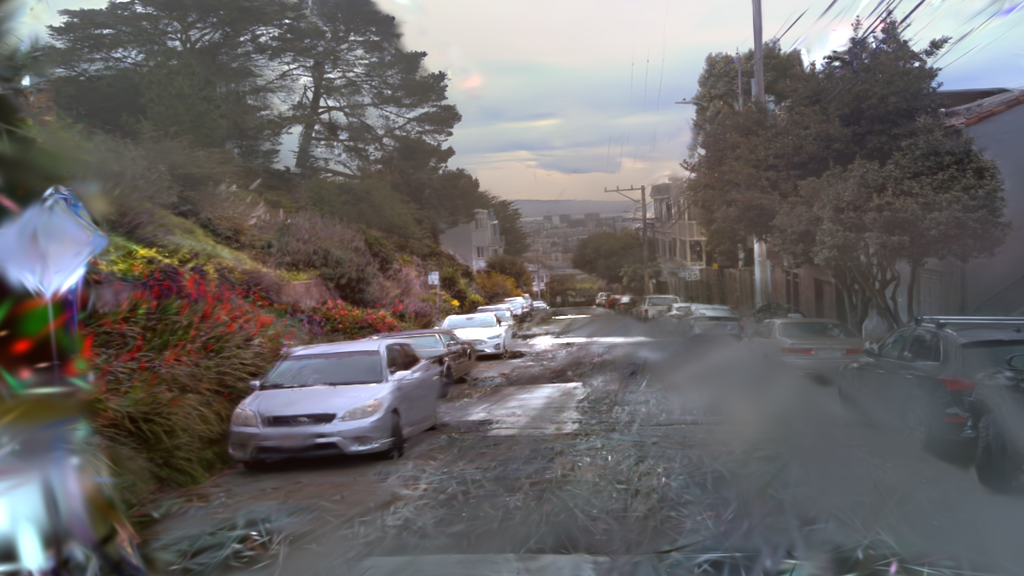}\end{tabular} &
\begin{tabular}{l}\includegraphics[width=0.198\linewidth]{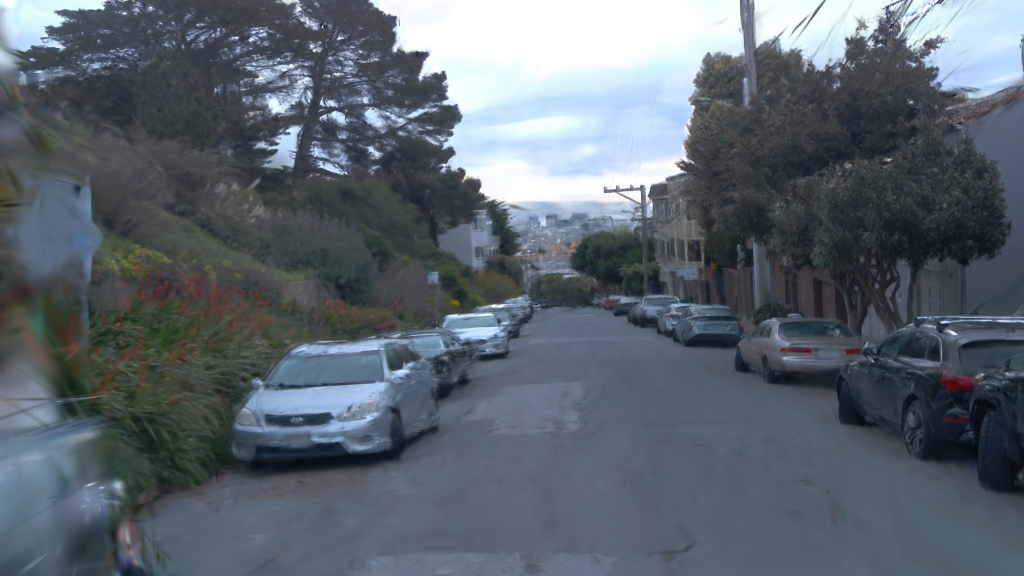}\end{tabular} &
\begin{tabular}{l}\includegraphics[width=0.198\linewidth]{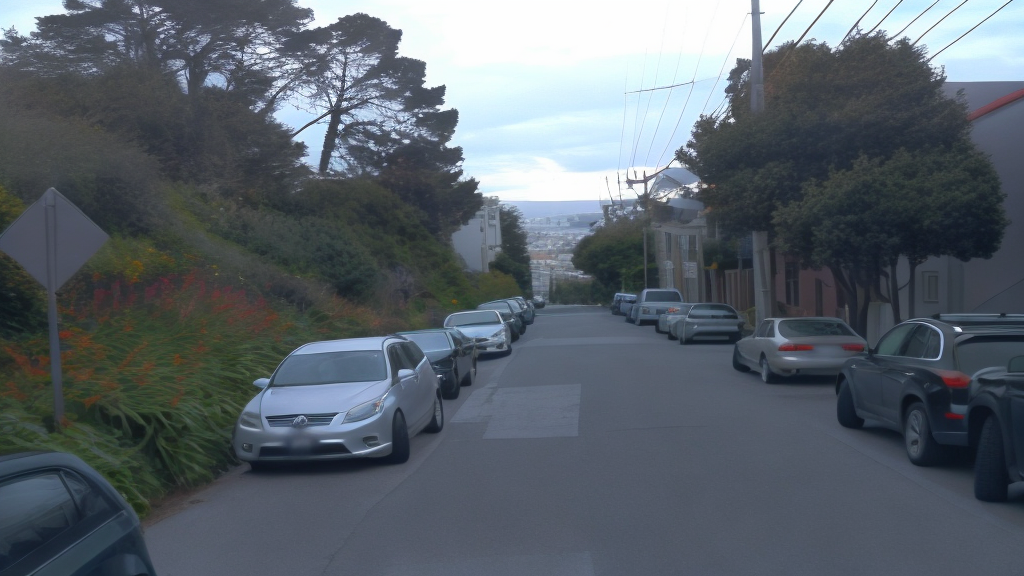}\end{tabular} &
\begin{tabular}{l}\includegraphics[width=0.198\linewidth]{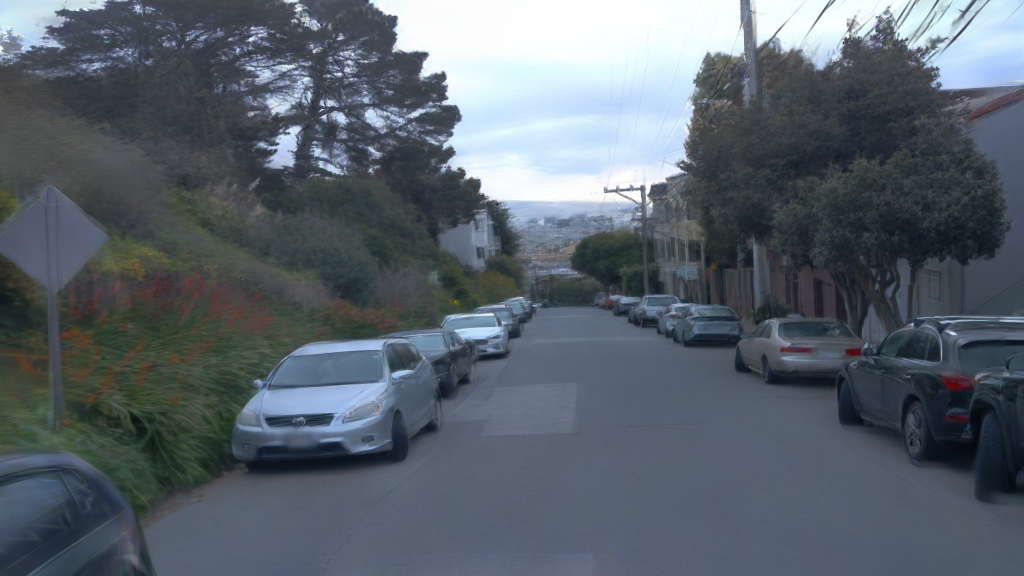}\end{tabular} \\

\end{tabular}}
\caption{
\textbf{Qualitative Comparison of Novel View Synthesis on the Open Waymo Dataset.}
We evaluate novel view synthesis using a 2-meter leftward camera offset. 
\textbf{Ours-R} refers to the outputs generated by our multi-modal diffusion model, which relies exclusively on sparse conditional inputs (color and depth). 
In contrast, \textbf{Ours-S} represents the results of the 3D Gaussian Splatting (3DGS) model, which is trained in a supervised manner using the outputs from Ours-R module as training signals. 
Our approach achieves photorealistic consistency and demonstrates superior geometric stability under significant viewpoint changes, outperforming existing baseline methods.
}
\label{fig:comparison}
\end{figure*}

We compare the visualization results of 3DGS~\cite{kerbl20233dgs}, S$^3$Gaussian~\cite{huang2024S3Gaussian}, and StreetGaussian~\cite{yan2024streetgs} with our model in Fig.~\ref{fig:comparison}, where our method generates higher-quality novel views under shifted viewpoints.
Notably, Ours-R demonstrates robust generalization with sparse input conditions, producing photorealistic lane-shift renderings without per-scene optimization. Meanwhile, Ours-S enhances overall consistency and geometric detail by integrating the 3D Gaussian Splatting model. This hybrid approach effectively balances the fidelity of diffusion-based generation with the explicit representation of 3D scenes, surpassing baseline methods in both visual realism and structural preservation under extreme viewpoint variations.

\begin{figure}[!t]
\centering
\includegraphics[width=\linewidth]{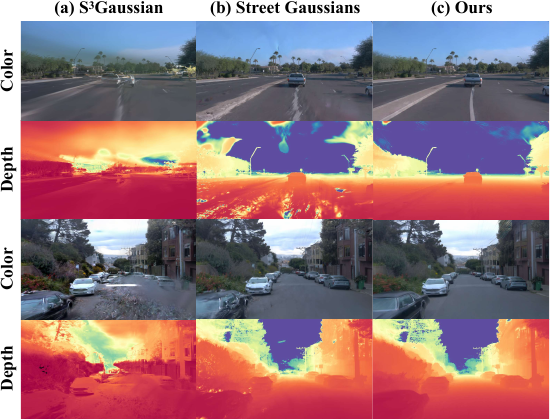}
\caption{
\textbf{The qualitative comparisons of depth rendering.}
The results demonstrate the high geometric quality of our method, achieving superior depth rendering outcomes.
}
\vspace{-0.3cm}
\label{fig:depth_vis}
\end{figure}

Fig.~\ref{fig:depth_vis} visualizes the depth maps in the novel viewpoints, demonstrating that our method, which combines multi-modal diffusion and 3DGS maintained the better geometric quality under significant viewpoint offsets. 

\subsubsection{Ablation Study}
In this section, we conduct the ablation study to analyze the effectiveness of each module of MuDG.

\begin{table}[!t]
\centering
\resizebox{\linewidth}{!}{
\begin{tabular}{l|cc|cccc}
\toprule
Setting    &   Ref. Img   &   Sparse Cond.     & SSIM $\uparrow$  &  PSNR $\uparrow$   &  LPIPS $\downarrow$ &  FID   $\downarrow$ \\
\midrule
(a)   & \usym{2713}  &   -    & 0.539 & 19.09 & 0.325 & 31.33 \\ 
(b)   & - &  \usym{2713}  & 0.631 & 22.01 & 0.254 & 25.17 \\ 
(c) & \usym{2713} & \usym{2713}  & \underline{0.642} & \underline{22.48}  & \underline{0.238} & \underline{23.66} \\ 
\midrule
\rowcolor{gray!10}Full Model  & \usym{2713} & \usym{2713}  & \textbf{0.643} & \textbf{22.49} & \textbf{0.237} & \textbf{23.29} \\
\bottomrule
\end{tabular}
}
\vspace{-5pt}
\caption{
\textbf{Ablation Studies on Control Signal.}
Module (a) utilizes only reference images as control signal, while module (b) employs sparse conditions. 
Module (c) simply concatenate both information, whereas our full model injects the reference image only into the first frame of the sparse condition while maintaining other sparse conditions as spatiotemporal control signals.
The results underscore the efficiency and quality of our designed model.
}
\vspace{-0.1cm}
\label{tab:mdm_ablation_results}
\end{table}

\begin{table}[!t]
\centering
\resizebox{0.99\linewidth}{!}{
\begin{tabular}{l|ccc|ccc|cc}
\toprule
Setting & $\mathbf{I}_{v}$ & $\mathbf{D}_{v}$ &  $\mathbf{S}_{v}$ & SSIM $\uparrow$  & PSNR $\uparrow$   & LPIPS $\downarrow$  & FID $\downarrow$    & FVD  $\downarrow$    \\
\midrule
(a) &  -  &   - & - & \textbf{0.916} & \textbf{31.96} & \textbf{0.085} & 39.62 & 283.05 \\
(b) & \usym{2713}  &   - & - & \underline{0.910} & 31.46 & 0.091 & 32.20 & 217.17 \\
(c) & \usym{2713}  &   \usym{2713} & - & {0.910} & 31.54 & {0.090} & \underline{32.00} & 216.14 \\
(d) & \usym{2713}  &   - & \usym{2713} & 0.910 & 31.51 & 0.091 & 32.17 & \underline{214.22} \\
\midrule
\rowcolor{gray!10}Full Model &\usym{2713}  &   \usym{2713} & \usym{2713} & {0.910} & \underline{31.60} & \underline{0.090} & 
\textbf{31.94} & \textbf{209.79} \\
\bottomrule
\end{tabular}
}
\vspace{-5pt}
\caption{
\textbf{Ablation on Multi-modal Supervision for 3DGS.}
The results demonstrate the significance of multi-modal supervision for 3DGS reconstruction. 
All multi-modal supervisions contribute to the 3DGS representation, particularly in dynamic urban scenes.
}
\label{tab:3dgs_ablation_results}
\vspace{-10pt}
\end{table}

\noindent\textbf{Ablation on Control Signal.} 
As shown in Tab.~\ref{tab:mdm_ablation_results}, we evaluate our diffusion model’s novel view synthesis capability under different control signals.
The base model using only reference images (Tab.~\ref{tab:mdm_ablation_results} (a)) performs the weakest (SSIM 0.54, FID 31.33), highlighting the limitations of static single-view inputs.
Using only sparse conditions (Tab.~\ref{tab:mdm_ablation_results} (b)) improves SSIM by 17\% and reduces FID by 19.6\%, demonstrating their effectiveness in encoding camera poses and motion trajectories.
However, sparse conditions alone remains insufficient for consistent generation in unobserved regions.
Combining reference images with sparse conditions (Tab.~\ref{tab:mdm_ablation_results} (c)) leads to moderate improvements: PSNR increases from 22.01 to 22.48, while LPIPS decreases by 6\%, indicating the complementary nature of static and dynamic cues.
To further enhance performance, our optimized architecture strategically places the reference image as the first sparse input frame (Full Model), achieving superior metrics while reducing channel capacity by half (from 8 to 4 channels).
This design not only preserves spatiotemporal consistency but also replace brute-force channel expansion via concatenation and mitigating feature misalignment.
Additionally, our streamlined architecture reduces network parameters, improving efficiency without compromising synthesis quality.

\noindent\textbf{Ablation on Multi-modal Supervision for 3DGS.} 
The Tab.~\ref{tab:3dgs_ablation_results} quantifies the impact of virtual viewpoint multi-modal results (dense color $\mathbf{I}_{v}$, depth $\mathbf{D}_{v}$, and semantic maps $\mathbf{S}_{v}$) generated from MDM serving as supervision on 3DGS reconstruction. 
Integrating multi-modal virtual views significantly improves the quality of the novel view synthesis. 
Specifically, incorporating virtual view color information reduces FID by 18.7\% and improves FVD by 23.3\%, indicating that augmented virtual views effectively bridge the distribution gap between synthetic and real data while improving temporal coherence in video sequences.
Additionally, depth information from virtual views enhances geometric consistency, while semantic information strengthens scene structural coherence. Although virtual views introduce minor pixel-level deviations (SSIM / PSNR drop $\sim$0.5\%), multimodal fusion significantly improves scene plausibility and temporal stability, as reflected in a progressive 26\% FVD reduction. These findings highlight the critical role of cross-view alignment in dynamic scene modeling, particularly for urban scene view synthesis.

\vspace{-0.1cm}

\section{Conclusion}
\label{sec:conclusion}

We present MuDG, an innovative framework that synergizes controllable multi-modal diffusion models with 3D Gaussian Splatting for robust urban scene reconstruction and view synthesis. By conditioning our diffusion model on aggregated LiDAR data and multi-modal priors, our proposed MDM is able to synthesize multi-modal results under novel viewpoints with no need of per-scene optimization. The synthesized dense RGB, depth, and semantic outputs not only enable feed-forward novel view synthesis but also significantly improve 3DGS training through enriched supervision signals. Extensive experiments validate that our framework achieves state-of-the-art performance in urban scene reconstruction and novel view synthesis.
\vspace{-0.3cm}
\noindent\paragraph{Limitations and Future Work.}
Our MDM is built on a denoising diffusion UNet, which entails substantial time costs for both training and inference. Thus, accelerating the model is a key area for improvement. Additionally, our training process relies on pre-processed pseudo ground truths (depth and semantic maps), further contributing to the overall time expense. Optimizing the generation process of pseudo GTs will be another crucial focus for future enhancements.
{
    \small
    \bibliographystyle{ieeenat_fullname}
    \bibliography{main}
}

\end{document}